\documentclass[10pts,twocolumn,letterpaper]{article}

\usepackage{cvpr}
\usepackage{times}
\usepackage{epsfig}
\usepackage{graphicx}
\usepackage{amsmath}
\usepackage{amsthm}
\usepackage{algorithm}
\usepackage{algorithmic}
\usepackage{amssymb}
\usepackage{subfigure}
\usepackage{dsfont}
\usepackage{float}
\usepackage{booktabs}
\usepackage{soul}
\usepackage{wrapfig}
\usepackage[usenames,dvipsnames]{color}
\usepackage{arydshln}
\usepackage{enumitem}
\usepackage[table]{xcolor}
\usepackage{multirow}
\usepackage{tabularx}
\usepackage{setspace}

\usepackage[pagebackref=false,breaklinks=true,letterpaper=true,colorlinks,bookmarks=false]{hyperref}

\DeclareMathOperator*{\argmin}{\arg\!\min}
\graphicspath{{images/}}
\newcommand{\comment}[1]{}

\newcommand{\vincentrmk}[1]{\textcolor{green}{\bf vl: {#1}}}

\newcommand{\pascalrmk}[1]{\textcolor{red}{\bf pf: #1}}

\newcommand{\bugrarmk}[1]{\textcolor{blue}{\bf bt: #1}}

\newcommand{\artemrmk}[1]{{\color{violet} \bf ar: #1}}
\newcommand{\xiaolurmk}[1]{\textcolor{cyan}{\bf xl: #1}}

\comment{
\newcommand{\vincentrmk}[1]{}

\newcommand{\pascalrmk}[1]{}

\newcommand{\bugrarmk}[1]{}

\newcommand{\artemrmk}[1]{}
\newcommand{\xiaolurmk}[1]{}

}

\newcommand{\bI}{\mathbf{I}}
\newcommand{\bX}{\mathbf{X}}

\newcommand{\bY}{\mathbf{Y}}
\newcommand{\bZ}{\mathbf{Z}}

\newcommand{\bm}{\mathbf{m}}

\cvprfinalcopy 

\def\cvprPaperID{166} 

\ifcvprfinal\fi
\begin{document}


\title{Direct Prediction of 3D Body Poses from Motion Compensated Sequences}


\author{Bugra Tekin$^1$ \quad\quad Artem Rozantsev$^{1}$ \quad\quad Vincent Lepetit$^{1,2}$ \quad\quad Pascal Fua$^1$ \\
$^1$CVLab, EPFL, Lausanne, Switzerland, {\tt\small \{firstname.lastname\}@epfl.ch}\\
$^2$TU Graz, Graz, Austria, {\tt\small lepetit@icg.tugraz.at}}

\maketitle


\begin{abstract}

We propose an efficient approach to exploiting motion information from consecutive frames of a video sequence to recover the 3D pose of people.  Previous approaches typically compute candidate poses in individual frames and then link them in a post-processing step to resolve ambiguities. By contrast, we directly regress from a spatio-temporal volume of bounding boxes to a 3D pose in the central frame. 

We further show that, for this approach to  achieve its full potential, it is essential to compensate for the motion in consecutive frames so that the subject remains centered. This then allows us to effectively overcome  ambiguities and improve upon the state-of-the-art by a large margin on the Human3.6m, HumanEva, and KTH Multiview Football 3D human pose  estimation benchmarks.
\end{abstract}





\section{Introduction}
\label{sec:intro}

In recent years, impressive motion  capture results have been demonstrated using
depth cameras, but 3D body pose  recovery from ordinary monocular video sequences 
remains extremely challenging. Nevertheless, there is  great interest in doing so, 
both because cameras are  becoming ever cheaper and more prevalent  and  because 
there are many  potential   applications.   These  include   athletic   training,
surveillance, and entertainment.

Early approaches to monocular 3D pose tracking involved recursive frame-to-frame
tracking and were  found to be brittle, due to  distractions and occlusions from
other people or objects    in   the  scene~\cite{Urtasun05b}.  Since   then,  the
focus has shifted  to ``tracking by detection,'' which  involves detecting human
pose more  or less independently  in every  frame followed by  linking the poses
across the frames~\cite{Andriluka10,Ramanan05},  which  is   much  more  robust  to
algorithmic   failures  in   isolated  frames.   More  recently,   an  effective
single-frame approach to learning a regressor  from a kernel embedding of 2D HOG
features  to 3D  poses has  been proposed~\cite{Ionescu14a}.   Excellent results
have also been reported using a Convolutional Neural Net~\cite{Li14a}.

However,  inherent  ambiguities of  the  projection  from  3D to  2D,  including
self-occlusion  and   mirroring,  can   still  confuse   these  state-of-the-art
approaches. A linking  procedure can correct for these ambiguities  to a limited
extent  by  exploiting  motion  information  \emph{a  posteriori}  to  eliminate
erroneous  poses by  selecting  compatible candidates  over consecutive  frames.
However,  when such  errors  happen  frequently for  several  frames  in a  row,
enforcing temporal consistency afterwards is not enough.

\begin{figure}[t]
\vspace{3mm}
\centering
\begin{tabular}{cccc}
\hspace{-0.20cm}\includegraphics[width=0.22\linewidth, height=1.0in]{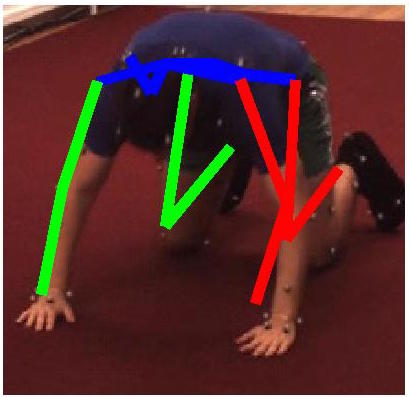}
&\hspace{-0.35cm}\includegraphics[width=0.22\linewidth, height=1.0in]{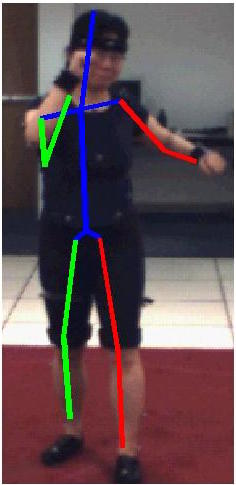}  
&\hspace{-0.35cm}\includegraphics[width=0.22\linewidth, height=1.0in]{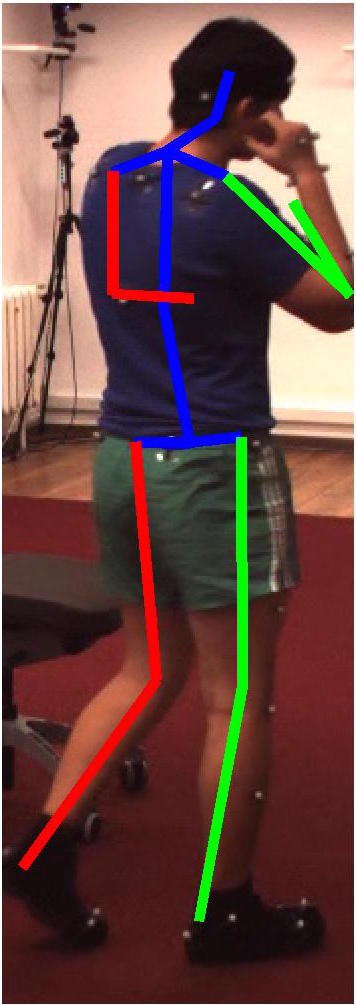}  
&\hspace{-0.35cm}\includegraphics[width=0.22\linewidth, height=1.0in]{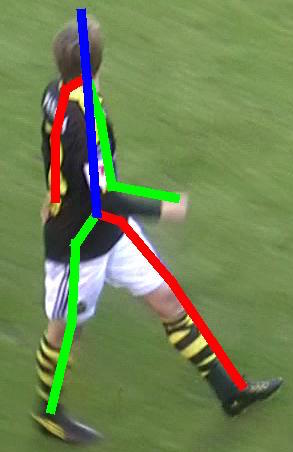}   \\
\hspace{-0.20cm}\includegraphics[width=0.22\linewidth, height=0.9in]{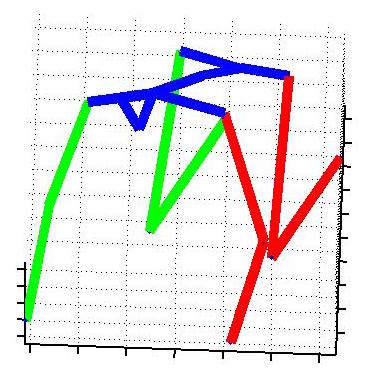}  
&\hspace{-0.35cm}\includegraphics[width=0.22\linewidth, height=1.0in]{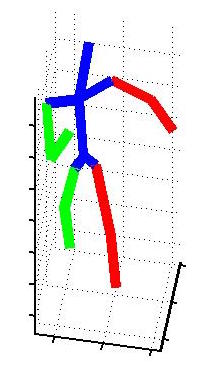}  
&\hspace{-0.35cm}\includegraphics[width=0.22\linewidth, height=0.94in]{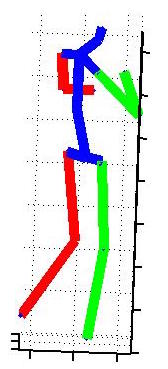} 
&\hspace{-0.35cm}\includegraphics[width=0.22\linewidth, height=0.95in]{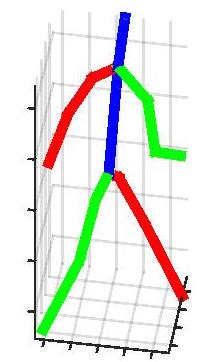} \\
\end{tabular}
\vspace{0.1cm}
\caption{3D human pose estimation  in Human3.6m, HumanEva and KTH Multiview Football datasets. 
The recovered 3D  skeletons are reprojected into the images
  in the top row and  shown by  themselves in the bottom row. Our approach can
  reliably recover  3D poses in complex scenarios by collecting appearance and motion
  evidence simultaneously from motion compensated sequences.  All the figures in this paper are best viewed in color.}
\vspace{-0.2cm}
\label{fig:intro}
\end{figure}

In  this paper,  we therefore  propose to  exploit motion  information from  the
start. To this end, we learn a regression function that directly predicts the 3D
pose in a  given frame of a  sequence from a spatio-temporal  volume centered on
it. This volume comprises bounding  boxes surrounding the person in consecutive
frames coming before and after the central one.  We will show that this approach
is more  effective than  relying  on   regularizing   initial  estimates \emph{a
posteriori}.  We  evaluated   different  regression schemes and obtained the best 
results by applying a Deep Network to the spatiotemporal features~\cite{Klaser08,Weinland10} 
extracted from the image volume.  Furthermore,  we show that,  for  this approach  
to perform to its best, it is  essential  to  align the  successive  bounding  
boxes of  the spatio-temporal volume so that  the  person  inside  them  remains  
centered. To this end, we trained two Convolutional  Neural Networks to  first 
predict large body shifts between consecutive frames and then refine them. This 
approach to motion compensation outperforms other more standard ones~\cite{Park13} 
and improves 3D human pose estimation accuracy significantly. Fig.~\ref{fig:intro}
depicts sample results of our approach.

The novel contribution of this paper is therefore a principled approach to 
combining appearance  and motion  cues  to  predict  3D  body pose  in  a
discriminative manner. Furthermore, we demonstrate that what  makes this
approach both practical  and effective is the compensation for the body motion
in consecutive frames of the spatiotemporal volume.  We  show that the  proposed
framework improves upon  the state-of-the-art~\cite{Andriluka10,Belagiannis14a,Bo10,Ionescu14a,Li14a}  
by   a large   margin  on  Human3.6m~\cite{Ionescu14a}, HumanEva~\cite{Sigal10}, 
and KTH Multiview Football~\cite{Burenius13} 3D human pose estimation benchmarks.


\section{Related Work}
\label{sec:related}

Approaches  to estimating  the 3D  human pose  can be  classified into  two main
categories, depending on  whether they rely on still images  or image sequences.
We briefly review both kinds below.  In the results section, we will demonstrate
that  we  outperform  state-of-the-art  representatives of  each  of  these  two
categories.

{\bf 3D Human Pose Estimation in Single Images.} Early approaches tended to rely
on generative models to search the  state space for a plausible configuration of
the      skeleton      that      would       align      with      the      image
evidence~\cite{Gall10,Gammeter08,Ormoneit00,Sidenbladh00}.  These methods remain
competitive provided  that a good  enough initialization can be  supplied.  More
recent  ones~\cite{Belagiannis14a,Burenius13}  extend   2D  pictorial  structure
approaches~\cite{Felzenszwalb10}  to the  3D  domain.  However,  in addition  to
their high computational cost, they  tend to have difficulty localizing people's
arms accurately  because the corresponding  appearance cues are weak  and easily
confused with the background~\cite{Sapp10}.

By             contrast,             discriminative             regression-based
approaches~\cite{Agarwal04a,Bo10,Ionescu14b,Sminchisescu05}   build   a   direct
mapping from image evidence to 3D  poses. Discriminative methods have been shown
to   be   effective,   especially   if    a   large   training   dataset,   such
as~\cite{Ionescu14a} is available.  Within  this context, rich features encoding
depth~\cite{Shotton11}  and  body part  information~\cite{Ionescu14b,Li14a}  have
been shown to be effective at increasing the estimation accuracy. However, these
methods can still suffer from  ambiguities such as self-occlusion, mirroring and
foreshortening, as  they rely on  single images.   To overcome these  issues, we
show how to use not only appearance, but also motion features for discriminative
3D human pose estimation purposes.
 
In  another  notable study,  \cite{Bo10}  investigates  merging image features  across
multiple views.   Our method  is fundamentally  different as we  do not  rely on
multiple cameras. Furthermore, we compensate for apparent motion of the person's
body  before  collecting  appearance  and motion  information  from  consecutive
frames.

{\bf 3D  Human Pose  Estimation in  Image Sequences.} Such approaches also fall 
into two main classes.

The    first   class    involves   frame-to-frame    tracking   and    dynamical
models~\cite{Urtasun05b} that  rely on Markov dependencies  on previous  frames.  
Their main  weakness is that they require  initialization and cannot recover from 
tracking failures.

To address these  shortcomings, the second class focuses  on detecting candidate
poses  in  individual  frames  followed  by linking  them  across  frames  in  a
temporally consistent manner.  For  example, in~\cite{Andriluka10}, initial pose
estimates are refined using 2D tracklet-based estimates. In~\cite{Zuffi13}, dense  
optical   flow  is  used   to  link articulated shape  models in  adjacent frames. 
Non-maxima suppression  is then employed to merge pose estimates  across frames 
in~\cite{Burgos13}.  By contrast to these approaches, we capture the  temporal 
information earlier in the process by extracting  spatiotemporal features from  
image cubes of short  sequences and regressing  to  3D  poses.  Another 
approach~\cite{Brauer11} estimates  a  mapping from consecutive ground-truth 2D 
poses  to a central 3D  pose. Instead, we do not require any such 2D pose 
annotations and directly use as input a sequence of motion-compensated frames.

While       they      have       long      been       used      for       action
recognition~\cite{Laptev05b,Weinland10}, person  detection~\cite{Park13}, and 2D
pose  estimation~\cite{Ferrari08}, spatiotemporal  features have  been underused
for 3D body pose estimation purposes.  The  only recent approach we are aware of
is that  of~\cite{Zhou14a} that  involves building a  set of  point trajectories
corresponding to high joint responses and  matching them to motion capture data.
One drawback of  this approach is its very high  computational cost. Also, while
the 2D  results look promising, no  quantitative 3D results are  provided in the
paper and no code is available for comparison purposes.


\section{Method}
\label{sec:method}

\begin{figure*}[t!]

  \centering
  \includegraphics[width=0.98\textwidth, height=1.75in]{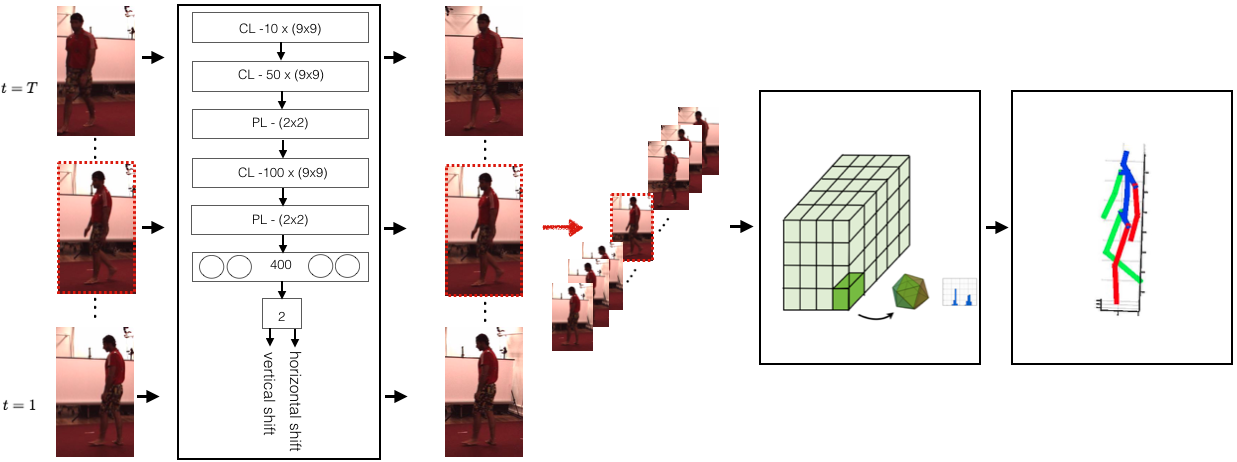}
  
  \footnotesize   \noindent  \hspace{-2mm}    (a)   Image    stack   \hspace{4mm}    (b)   Motion
  compensation \hspace{7mm}  (c) Rectified spatiotemporal volume  \hspace{1.5mm} (d)  
  Spatiotemporal feautures \hspace{3mm} (e) 3D pose regression \\[1mm] 
  
  \caption[Overview] {Overview of our approach  to 3D pose estimation. {\bf(a)} A  person  is  
  detected  in   several  consecutive  frames.   {\bf(b)}  Using a CNN, the corresponding image  
  windows are shifted  so that the  subject remains centered.  {\bf(c)} A \emph{rectified 
  spatiotemporal volume} (RSTV) is formed by concatenating the aligned windows.  {\bf(d)} A pyramid 
  of 3D  HOG features are extracted densely over the  volume.  {\bf(e)}  The 3D  pose in  the 
  central  frame is  obtained by regression.}
  
  \label{fig:approach}
  \vspace{-6mm}
\end{figure*}

Our  approach  involves finding  bounding  boxes  around people  in  consecutive
frames,  compensating for the motion   to  form spatiotemporal volumes, and 
learning a mapping from these volumes  to a 3D pose in their central frame. 

In the  remainder of  this section,  we first introduce  our formalism  and then
describe each individual step, depicted by Fig.~\ref{fig:approach}.

\subsection{Formalism}
\label{subsec:formalism}

In this work,  we represent 3D body  poses in terms of skeletons,  such as those
shown in Fig.~\ref{fig:intro}, and the 3D locations of their $D$ joints relative
to that of  a root node.  As several authors  before us~\cite{Bo10, Ionescu14a},
we chose this  representation because it is well adapted  to regression and does
not  require us  to  know {\it  a  priori}  the exact  body  proportions of  our
subjects. It  suffers from  not being orientation  invariant but  using temporal
information     provides     enough     evidence     to     overcome     this
difficulty.

Let $\bI_i$ be  the $i$-th image of  a sequence containing a  subject and $\bY_i
\in \mathbb{R}^{3\cdot D}$ be a vector  that encodes the corresponding 3D joint
locations. Typically, regression-based  discriminative  approaches to  inferring  $\bY_i$
involve  learning a parametric~\cite{Agarwal04a,Kanaujia07} or
non-parametric~\cite{Urtasun08} model of  the mapping function, $\bX_i
\rightarrow  \bY_i \approx  {\bf  f}({\bX_i})$ over  training  examples, where $\bX_i  = \Omega(\bI_i;
\bm_i)$ is  a feature vector  computed over the  bounding box or  the foreground
mask, $\bm_i$,   of    the    person   in    $\bI_i$.  The  model parameters are  
usually learned  from a  labeled set  of $N$  training examples,
$\mathcal{T} =  \{ (  \bX_i, \bY_i  )\}_{i=1}^N$.    As   discussed    in
Section~\ref{sec:related}, in such a setting, reliably estimating the 3D pose is
hard to do due  to the inherent ambiguities of 3D human  pose estimation such as
self-occlusion and mirror ambiguity.

Instead, we model the mapping function ${\bf f}$ conditioned on a spatiotemporal
3D data  volume consisting of  a sequence of $T$  frames centered at  image $i$,
${\bf V}_i = [ {\bf I}_{i - T/2 + 1},\dots, {\bf I}_{i}, \dots, {\bf I}_{i + T/2
}]$, that is, $\bZ_i  \rightarrow {\bY_i} \approx {\bf f}({\bZ_i})$ where  $ {\bf Z}_i =
\xi({\bf V}_i;{\bf m}_{i - T/2 + 1}, \dots, {\bf  m}_i, \dots, {\bf m}_{i +
  T/2 })$ is  a feature vector computed  over the data volume,  ${\bf V}_i$. The
training  set,  in this  case,  is  $\mathcal{T} =  \{  ({\bf  Z}_i, {\bf  Y}_i)
\}_{i=1}^N$, where  ${\bf Y}_i$ is  the pose in the  central frame of  the image
stack.  In practice, we collect every block of consecutive $T$ frames across all
training videos to obtain data volumes. We will show in the results section that
this significantly improves  performance and that the best  results are obtained
for volumes of $T=24$ to $48$ images, that is 0.5 to 1 second given the 50fps of
the sequences of the Human3.6m~\cite{Ionescu14a} dataset.

\vspace{-1mm}

\subsection{Spatiotemporal Features}
\label{subsec:features}

Our feature  vector $\bZ$  is  based on the 3D  HOG descriptor~\cite{Weinland10}, 
which {\it  simultaneously} encodes appearance and motion information. It is 
computed by first  subdividing a data volume such as the one  depicted by 
Fig.~\ref{fig:approach}(c) into  equally-spaced cells. For each one, the histogram
of   oriented  3D   spatio-temporal gradients~\cite{Klaser08} is then computed.   
To increase the descriptive power, we  use  a multi-scale  approach. We  compute
several 3D HOG  descriptors using different cell sizes. In practice, we use 3 levels
in the spatial dimensions---$2\times2$, $4\times4$  and $8\times8$---and  we set  the
temporal cell size  to a small  value---$4$ frames for  $50$ fps videos---to 
capture fine temporal  details.  Our  final feature  vector $\bZ$ is obtained by
concatenating the descriptors at multiple resolutions  into a  single vector.

An alternative  to encoding motion information in this way  would have been to
explicitly track body pose in the spatiotemporal volume,   as  done 
in~\cite{Andriluka10}. However, this involves detection of the body pose in
individual frames which is  subject to ambiguities caused  by the  projection
from  3D to  2D as explained in Section~\ref{sec:intro} and not having to do
this is a contributing factor to the good results we will show in
Section~\ref{sec:results}.

Another approach for spatiotemporal feature extraction could be to use  3D CNNs 
directly  operating   on  the  pixel intensities of  the spatiotemporal  volume.
However, in our experiments, we have observed that, 3D CNNs  did not achieve any
notable improvement in performance compared to spatial CNNs. This is likely due
to the  fact that 3D CNNs remain stuck in local minima due to the complexity of
the model and the large input dimensionality. This is also observed
in~\cite{Karpathy14,Mansimov15}.

\vspace{-1mm}


\subsection{Motion Compensation with CNNs}
\label{subsec:cnnmc}

\begin{figure}[t]
\centering
\begin{tabular}{c}
\includegraphics[width=0.20\linewidth, height=1in]{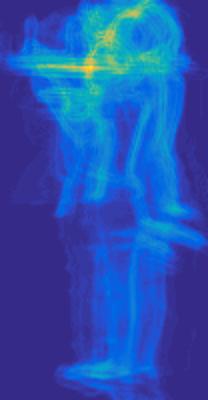} \hspace{5mm} 
\includegraphics[width=0.20\linewidth, height=1in]{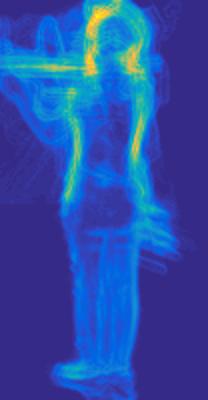}   \\
\footnotesize  (a) No compensation \hspace{15mm}   (b) Motion compensation \\
\end{tabular}
\caption[Motion stabilization]{Heat maps of the gradients across all frames for 
	\emph{Greeting} action {\bf (a)}  without and {\bf (b)} with  motion compensation. 
	When motion compensation is applied, body parts become covariant with the 3D HOG cells 
	across frames and thus the extracted spatiotemporal features become more part-centric 
	and stable.}
\vspace{-5mm}
\label{fig:motcomp}
\end{figure}

For the 3D HOG descriptors introduced above to be representative of the person's
pose, the  temporal bins must correspond  to specific body parts,  which implies
that the person should remain centered from frame to frame in the bounding boxes
used to  build the  image volume. We use  the Deformable  Part Model detector 
(DPM)~\cite{Felzenszwalb10} to  obtain these  bounding boxes,  as it proved to
be  effective in various applications.  However,  in practice, these bounding
boxes may not be well-aligned  on the person.  Therefore, we need to first shift
these boxes as shown in Fig.~\ref{fig:approach}(c) before creating a
spatiotemporal volume. In Fig.~\ref{fig:motcomp}, we illustrate this requirement
by showing heat maps of the gradients  across a sequence without and with motion
compensation.   Without it,  the gradients  are dispersed  across the  region of
interest, which reduces feature stability.

We therefore  implemented an  object-centric motion  compensation scheme
inspired  by  the  one  proposed in~\cite{Rozantsev15a}  for  drone  detection
purposes,  which  was   shown  to  perform  better   than  optical-flow  based
alignment~\cite{Park13}. To this end, we train regressors to estimate the shift
of the person from the center of the bounding box. We apply these shifts to the
frames of the image stack  so that the subject remains centered, and obtain what
we call a rectified spatio-temporal volume~(RSTV), as depicted in
Fig.~\ref{fig:approach}(c). We have chosen CNNs as our regressors, as they prove
to be effective in various regression tasks.

More formally, let  $m$  be  an image  patch  extracted from  a   bounding  box
returned  by DPM.   An ideal  regressor  $\psi(\cdot)$ for  our   purpose would
return the horizontal and vertical shifts $\delta u$ and $\delta   v$ of the
person from the center of $m$: $\psi(m) = (\delta u, \delta v)$.   In  practice,
to  make the  learning task  easier, we  introduce two separate regressors
$\psi_{coarse}(\cdot)$  and  $\psi_{fine}(\cdot)$.  We  train  the first one to
handle large  shifts and  the second  to refine  them.  We  use them iteratively
as  illustrated  by  Algorithm~\ref{alg:regression}.  After  each iteration, we
shift  the images  by the  computed amount  and estimate  a new shift. This 
process   typically   takes  only   4   iterations,  2   using
$\psi_{coarse}(\cdot)$ and 2 using $\psi_{fine}(\cdot)$.

\begin{figure}[!t]
  \centering
  \includegraphics[width = 3.4in]{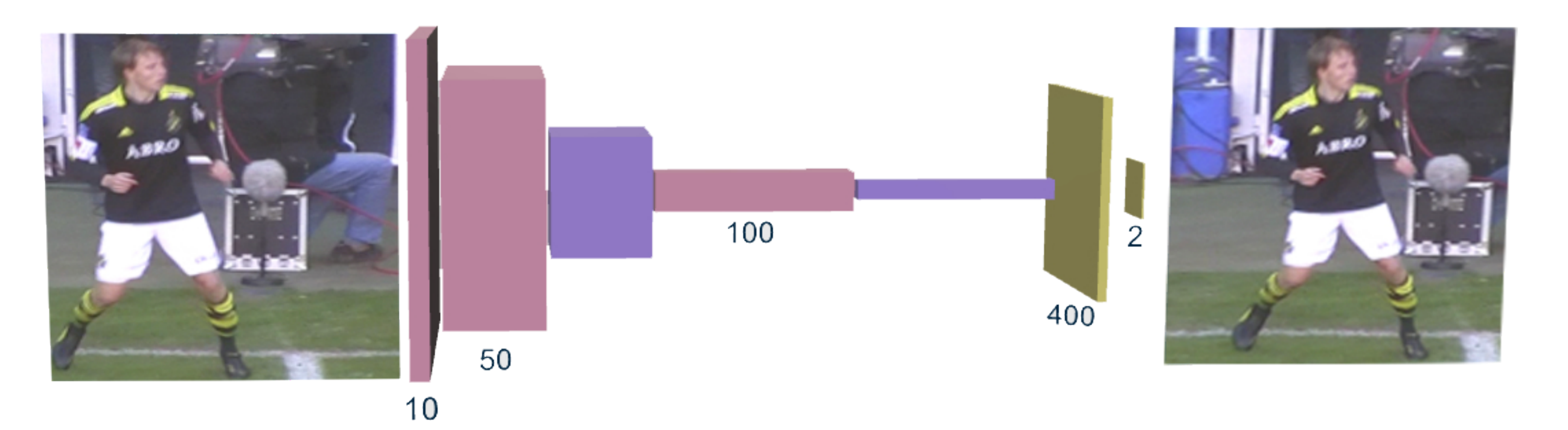}
  \caption{Motion Compensation CNN architecture. The network consists of convolution 
  	(dark red), pooling (purple) and fully connected (yellow) layers. The output of the 
  	network is a two-dimensional vector that describes horizontal and vertical shifts of 
  	the person from the center of the patch.}
  \label{fig:CNN_mc}
\end{figure}

Both CNNs feature the same  architecture,  which  comprises fully  connected,
convolutional,      and      pooling      layers,     as      depicted      by
Fig.~\ref{fig:approach}(b) and Fig.~\ref{fig:CNN_mc}. Pooling layers  are
usually used to  make the regressor robust  to small  image translations. 
However, while reducing the number of parameters to learn, they  could
negatively  impact performance as  our goal  is precise  localization. We 
therefore do  not use pooling  at the  first convolutional  layer, only  in the
subsequent ones. This yields accurate  results while keeping the number of 
parameters small enough to prevent overfitting.

\begin{algorithm}[!t]
\begin{spacing}{0.95}
{\fontsize{10}{10}\selectfont
    \begin{algorithmic}[0]
    \STATE {\bf Input}: image $I$, initial location estimate $(i,j)$
    \STATE       $\psi_{*}(\cdot)       =       \begin{cases}
      \psi_{coarse}(\cdot) \mbox{ for the first 2 iterations},
      \\ \psi_{fine}(\cdot) \mbox{ for the other 2}, \end{cases} $
    \STATE
    \STATE $(i^0,j^0) = (i,j)$
    \FOR {$o = 1:MaxIter$}
    \STATE $(\delta u^o,\delta v^o) = \psi_{*}(I(i^{o-1},j^{o-1}))$, \emph{with
    $I(i^{o-1},j^{o-1})$ the image patch in $I$ centered on $(i^{o-1},j^{o-1})$}
    \STATE $(i^o,j^o) = (i^{o-1}+\delta u^o,j^{o-1}+\delta v^o)$
    \ENDFOR
    \STATE $(i,j) = (i^{MaxIter},j^{MaxIter})$
  \end{algorithmic}}
  \end{spacing}
  \caption{Object-centric motion compensation.}
  \label{alg:regression}

\end{algorithm}

\vspace{1mm} {\bf Training  our  CNNs}  requires a set  of  image windows
centered on a subject, shifted versions, such as the one depicted by
Fig.~\ref{fig:CNN_mc}, and the corresponding shift amounts $(\delta u, \delta
v)$. We generate them from training data by randomly shifting ground truth
bounding boxes in horizontal and vertical directions. For $\psi_{coarse}$ these
shifts are large, whereas for $\psi_{fine}$ they are small, thus representing
the specific tasks of each regressor.

\vspace{1mm} {\bf Using  our CNNs} requires an initial  estimate of the bounding
box for every person, which  is given by DPM. However, applying the detector to
every frame of the  video is  time  consuming. Thus,  we decided to apply DPM
only to the first frame.  The position of the detection is  then refined  and 
the resulting  bounding  box is  used  as an  initial estimate in the second
frame. Similarly, its position is then corrected and the procedure is iterated
in subsequent frames. The initial person detector provides rough location
estimates and our motion compensation algorithm naturally compensates even for
relatively large positional inaccuracies using the regressor, $\psi_{coarse}$.
Some examples of our motion compensation algorithm, an analysis of its efficiency
as compared to optical-flow and further implementation details can be found in the supplementary material.

\subsection{Pose Regression}
\label{sec:regression}

We  cast 3D  pose estimation  in  terms of  finding a  mapping $\bZ  \rightarrow
\mathbf{f}(\bZ) \approx \bY$, where $\bZ$ is the 3D HOG descriptor computed over
a spatiotemporal volume and $\bY$ is the 3D pose in its central frame.  To learn
$\mathbf{f}$, we considered Kernel Ridge Regression (KRR)~\cite{Hofmann08}, Kernel
Dependency Estimation (KDE)~\cite{Cortes05} as they were used in
previous works on this task~\cite{Ionescu14b,Ionescu14a}, and Deep Networks.  

 {\bf Kernel  Ridge Regression (KRR)} trains  a model for each  dimension of the
 pose vector separately. To find the mapping from spatiotemporal features to 3D poses, it solves a 
 regularized least-squares problem of the form,
\vspace{-1mm}
\begin{equation}
\label{eq:kde_param}
\argmin_{\bf W} \sum_i || {\bf Y}_i - {\bf W} \Phi_Z({\bf Z}_i) ||_2^2 +  || {\bf W}||_2^2 \; \; ,
\vspace{-1mm}
\end{equation}

 \noindent where $(\bZ_j,\bY_j)$ are training  pairs and $\Phi_Z$ is the Fourier approximation to
the exponential-$\chi^2$ kernel~\cite{Ionescu14a}. This problem can be solved in closed-form by 
 ${\bf W} = (\Phi_Z({\bf Z})^T\Phi_Z({\bf Z})+ {\bf I})^{-1}\Phi_Z({\bf Z})^T\bY$.

\vspace{1mm}{\bf Kernel Dependency Estimation (KDE)} is a structured regressor
that accounts for correlations in  3D pose space. To  learn the  regressor, not
only the input as in the case of KRR, but also the  output vectors  are lifted
into high-dimensional   Hilbert  spaces   using kernel mappings   $\Phi_Z$  and 
$\Phi_Y$, respectively~\cite{Cortes05,Ionescu14a}. The dependency between high
dimensional input  and output spaces is modeled as a linear function.   The
corresponding  matrix ${\bf W}$  is computed by standard kernel ridge
regression,
\begin{equation}
\vspace{-1mm}
\label{eq:kde_param}
\argmin_{\bf W} \sum_i || \Phi_Y({\bf Y}_i) - {\bf W} \Phi_Z({\bf Z}_i) ||_2^2 +  || {\bf W}||_2^2 \; \; ,
\vspace{-1mm}
\end{equation}
To  produce  the  final prediction ${\bf Y}$, the  difference between the
predictions and  the mapping of the output in the high dimensional Hilbert space
is minimized by finding
\begin{equation}
\vspace{-1mm}
\label{eq:kde_sol}
\hat{\bY} = \argmin_\bY || {\bf W}^T \Phi_Z({\bf Z}) - \Phi_Y({\bf Y}) ||_2^2 \; \; .
\vspace{-1mm}
\end{equation}

Although  the problem  is  non-linear  and non-convex,  it  can nevertheless  be
accurately solved given the KRR  predictors for individual outputs to initialize
the process. In practice, we use an input kernel embedding based on
15,000-dimensional random feature maps  corresponding to  an
exponential-$\chi^2$ kernel, a 4000-dimensional output embedding corresponding 
to radial basis function kernel as  in~\cite{Li12b}.

\vspace{1mm}{\bf Deep Networks (DN)} rely on a multilayered architecture to
estimate the mapping to 3D poses.  We use 3 fully-connected  layers
with the
rectified linear unit (ReLU) activation function  in the first 2 layers and a
linear  activation function in the last layer. The first two layers consist of
3000 neurons each and the final layer has 51 outputs, corresponding to 17  3D 
joint  positions. We performed cross-validations across the network's
hyperparameters and choose the ones with the best performance on a validation set. We
minimize  the  squared  difference between  the prediction  and the
ground-truth  3D   positions  to  find  the mapping $\mathbf{f}$
parametrized by $\Theta$:
\vspace{-1mm}
\begin{equation}
\label{eq:nn_cost}
\hat{\Theta} = \argmin_{\Theta} \sum_{i} ||
\mathbf{f}_{\Theta}(\bZ_i) - \bY_i ||_2^2 \;\; .
\end{equation}

We used the ADAM~\cite{Kingma15} gradient update method to
steer  the optimization  problem  with  a learning  rate  of  0.001 and  dropout
regularization to prevent overfitting.  We will show in the results section that
our DN-based regressor outperforms KRR and KDE~\cite{Ionescu14b, Ionescu14a}. 
 

\section{Results}
\label{sec:results}

We   evaluate   our    approach   on the Human3.6m~\cite{Ionescu14a},
HumanEva-I/II~\cite{Sigal10}, and KTH  Multiview Football II~\cite{Burenius13}
datasets.  \emph{Human3.6m} is a  recently  released large-scale motion capture
dataset that  comprises 3.6 million  images and corresponding 3D  poses within
complex motion scenarios.  $11$ subjects perform $15$  different actions under
$4$ different viewpoints. In Human3.6m, different people appear in the
training and test data. Furhtermore, the data exhibits large  variations in terms
of  body shapes, clothing, poses and viewing angles within and  across
training/test splits~\cite{Ionescu14a}. The \emph{HumanEva-I/II} datasets  provide synchronized
images and  motion capture data and  are standard benchmarks for  3D human  pose
estimation. We further provide results on the \emph{KTH Multiview Football II}
dataset to demonstrate the performance  of our method in  a non-studio
environment. In this dataset, the cameraman follows the players as they move
around the pitch. We compare our method against several state-of-the-art
algorithms in these datasets. We chose them  to be representative  of different
approaches to 3D human pose  estimation, as discussed in
Section~\ref{sec:related}. For  those which we  do not  have access  to the
code, we used  the published performance numbers  and ran our own  method on the
corresponding data.

\begin{table*}[tbph]
	\begin{center}
		
		\tabcolsep=0.1cm
		
		\scalebox{0.75}{
			\begin{tabular}[b]{lccccccccccccccc}
				\toprule
				Method							        &Directions          &Discussion          &Eating             &Greeting            &Phone Talk         &Posing             &Buying             &Sitting\\ \toprule
				$e^{\chi_2}$-HOG+KRR~\cite{Ionescu14a}  &140.00 (42.55)      &189.36 (94.79)      &157.20 (54.88)     &167.65 (60.16)      &173.72 (60.93)     &159.25 (52.47)     &214.83 (86.36)     &193.81 (69.29) \\
				$e^{\chi_2}$-HOG+KDE~\cite{Ionescu14a}  &132.71 (61.78)      &183.55 (121.71)     &132.37 (90.31)     &164.39 (91.51)      &162.12 (83.98)     &150.61 (93.56)     &171.31 (141.76)    &151.57(93.84) \\				
				CNN-Regression~\cite{Li14a}			    &-	 	             &148.79 (100.49)     &104.01 (39.20)     &127.17 (51.10)      &-	               &-	               &-                  &-\\ 
				\midrule 
				RSTV+KRR (Ours)     					&119.73	(37.43)      &159.82 (91.81)      &113.42 (50.91)     &144.24 (55.94)      &145.62 (57.78)     &136.43 (44.49)     &166.01 (69.94)     &178.93 (69.32)\\ 
				RSTV+KDE (Ours)                         &103.32 (55.29)      &158.76 (119.16)     &89.22 (37.45)      &127.12 (76.58)      &119.35 (53.53)     &115.14 (65.21)     &{\bf108.12} (84.10)&{\bf136.82} (91.25)\\ 
				RSTV+DN (Ours)                          &{\bf102.41} (36.13) &{\bf147.72} (90.32) &{\bf88.83} (32.13) &{\bf125.28} (51.78) &{\bf118.02} (51.23)&{\bf112.38} (42.71)&129.17 (65.93)     &138.89 (66.18)\\ 				
				\bottomrule 			
				\toprule 	
				Method:									&Sitting Down        &Smoking             &Taking Photo       &Waiting             &Walking 		   &Walking Dog 	   &Walking Pair       & Average\\ 
				\toprule 			
				$e^{\chi_2}$-HOG+KRR~\cite{Ionescu14a}  &279.07 (102.81)     &169.59 (60.97)      &211.31 (83.72)     &174.27 (82.99)      &108.37 (30.63)     &192.26 (90.63)	   &139.76 (38.86)     &178.03 (67.47)\\ 
				$e^{\chi_2}$-HOG+KDE~\cite{Ionescu14a}  &243.03 (173.51)     &162.14 (91.08)      &205.94 (111.28)    &170.69 (96.38)      &96.60 (40.61)      &177.13(130.09)	   &127.88 (69.35)     &162.14 (99.38)\\ 
				CNN-Regression~\cite{Li14a}			   	&-	 	             &- 		          &189.08 (93.99)     &-       		       &77.60 (23.54)      &146.59 (75.38)	   &-      	           &-\\
				\midrule 			
				RSTV+KRR (Ours)		       				&247.21 (101.14)     &140.54 (56.04)      &192.75 (84.85)     &156.84 (78.13)      &70.98 (22.69)	   &152.01 (76.16)	   &91.47 (26.30)      &147.73 (61.52)\\ 
				RSTV+KDE (Ours)			 			    &{\bf206.43} (163.55)&119.64 (69.67)      &185.96 (116.29)    &146.91 (98.81)      &66.40 (20.92)      &128.29 (95.34)     &78.01 (28.70)	   &126.03 (78.39)\\ 
				RSTV+DN (Ours)			     		    &224.9 (100.63)      &{\bf118.42} (54.28) &{\bf182.73} (80.04)&{\bf138.75} (77.24) &{\bf55.07} (18.95) &{\bf126.29} (73.89)&{\bf65.76} (24.41) &{\bf124.97} (57.72)\\ 
				\bottomrule 
			\end{tabular} 
			}
	\end{center}
	\vspace{-2mm}
	\caption{3D  joint position  errors in \emph{Human3.6m}  using the metric of  average Euclidean distance between the  ground truth and predicted joint positions (in mm) to compare  our results,  obtained with  the 
		different regressors described in Section~\ref{sec:regression}, as well as for those of~\cite{Ionescu14a} and \cite{Li14a}. Our method achieves significant improvement over state-of-the-art discriminative regression 
		approaches by exploiting appearance and motion cues from motion compensated sequences. `-' indicates that the results are not reported for the corresponding action class. Standard deviations are given in parantheses.} 
	\vspace{-4mm}
	\label{tab:overall}
\end{table*}

\subsection{Evaluation on Human3.6m}
\label{subsec:human36m_eval}

To quantitatively  evaluate the performance of  our approach, we first  used the
recently  released Human3.6m~\cite{Ionescu14a}  dataset.  On  this dataset,  the
regression-based method of~\cite{Ionescu14a}  performed best at the  time and we
therefore use it  as a baseline.  That method relies  on a Fourier approximation
of 2D HOG features using the $\chi^2$ comparison metric, and we will refer to it
as ``$e^{\chi_2}$-HOG+KRR'' or ``$e^{\chi_2}$-HOG+KDE'', depending on whether it
uses KRR or KDE.  Since then, even better results have been obtained for some of
the actions by using CNNs~\cite{Li14a}. We denote it as CNN-Regression.  We refer
to our method as ``RSTV+KRR'', ``RSTV+KDE'' or ``RSTV+DN'', depending on whether
we use  respectively KRR, KDE, or  deep networks on the  features extracted from
the Rectified Spatiotemporal Volumes~(RSTV).  We report pose estimation accuracy
in terms  of average Euclidean  distance between the ground-truth  and predicted
joint positions~(in  millimeters) as  in~\cite{Ionescu14a,Li14a} and  exclude the
first and last $T/2$ frames~($0.24$ seconds for $T=24$ at 50 fps).

\newcommand{\mht}{0.93in}

The authors of~\cite{Li14a} reported results on  subjects S9 and S11 of Human3.6m
and  those  of~\cite{Ionescu14a} made  their  code  available.  To  compare  our
results to  both of  those baselines,  we therefore  trained our  regressors and
those of~\cite{Ionescu14a} for 15 different actions. We used 5 subjects~(S1, S5,
S6, S7, S8) for training purposes and  2~(S9 and S11) for testing.  Training and
testing  is  carried out  in  all  camera views  for  each  separate action,  as
described in~\cite{Ionescu14a}. Recall  from Section~\ref{subsec:formalism} that
3D body poses are represented by skeletons with $17$ joints.  Their 3D locations
are expressed relative  to that of a  root node in the coordinate  system of the
camera that captured the images.

\begin{figure*}[t]
	\centering
	\scalebox{0.92}{
		\begin{tabular}{cccccc}
			\includegraphics[width=0.12\linewidth, height=\mht]{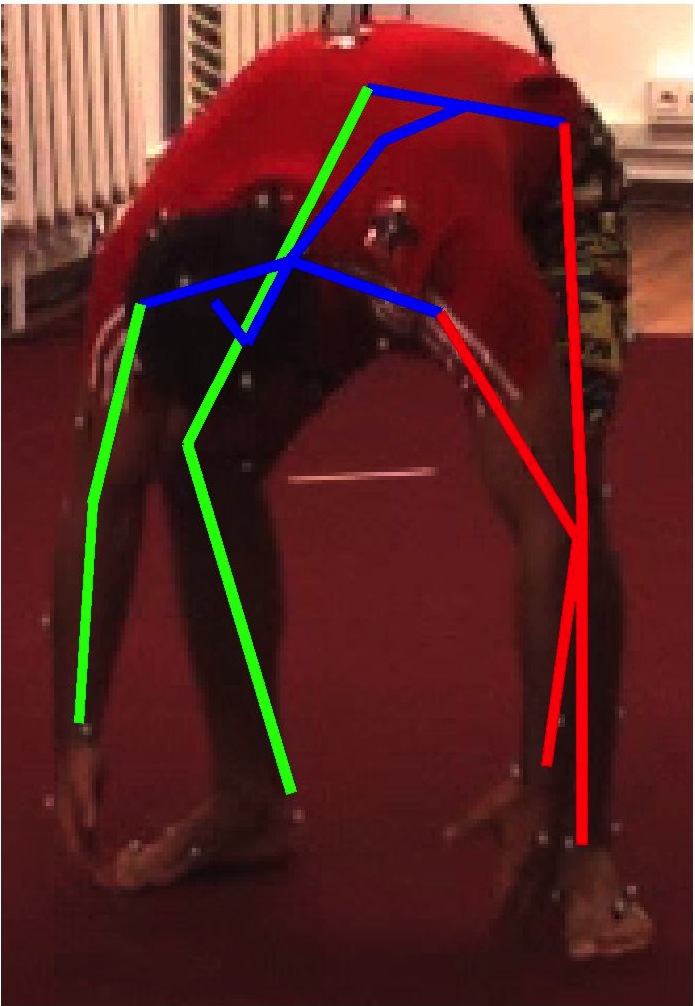} 
			&\includegraphics[width=0.13\linewidth, height=\mht]{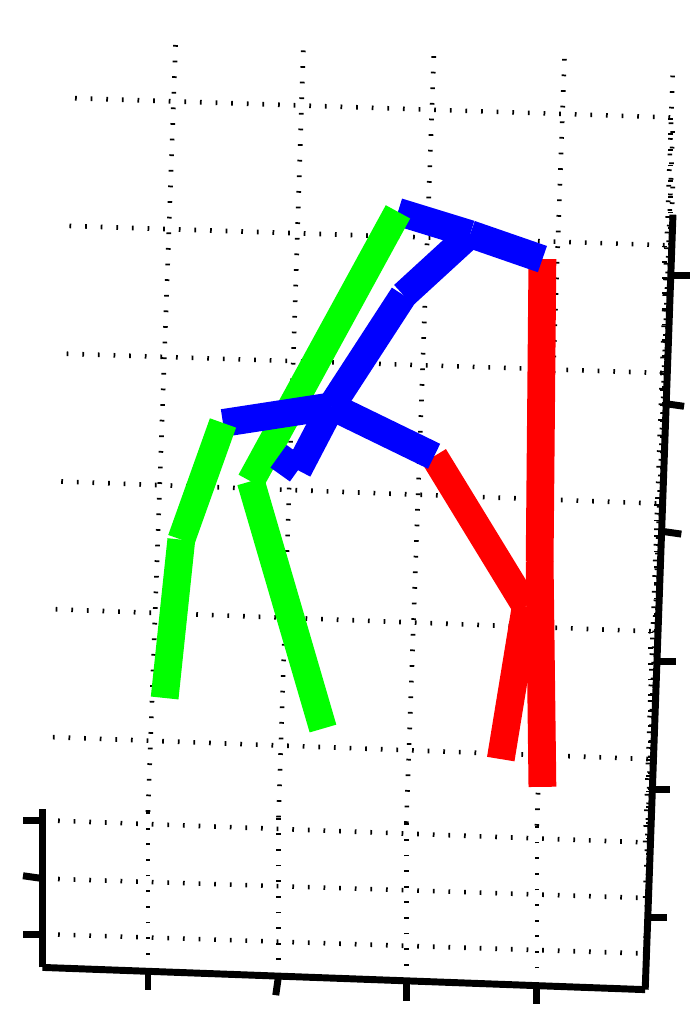}  
			&\includegraphics[width=0.12\linewidth, height=\mht]{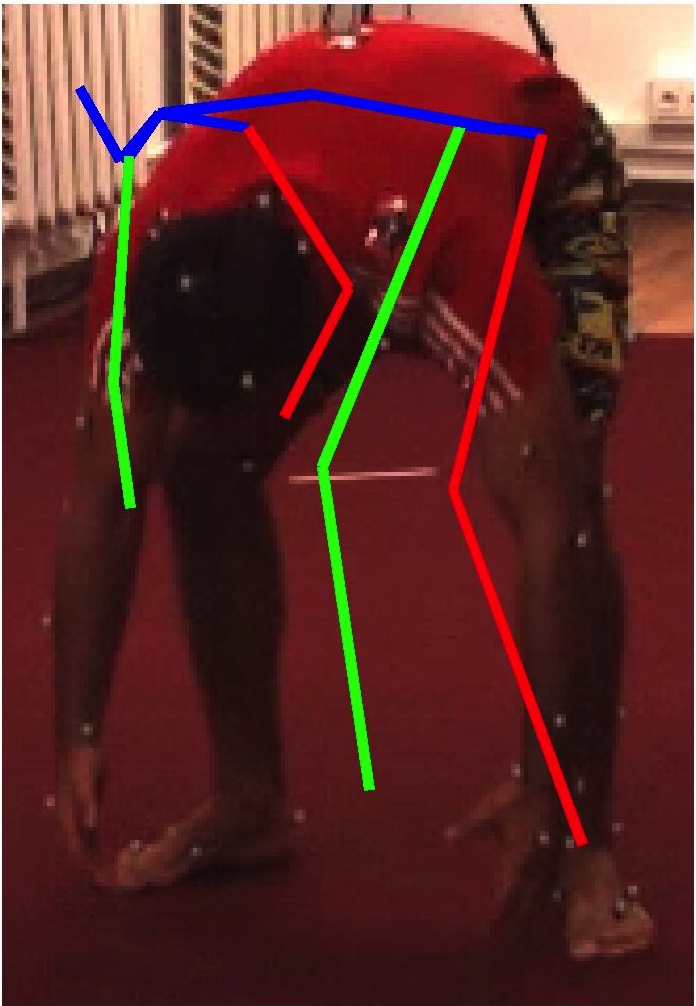}  
			&\includegraphics[width=0.13\linewidth, height=\mht]{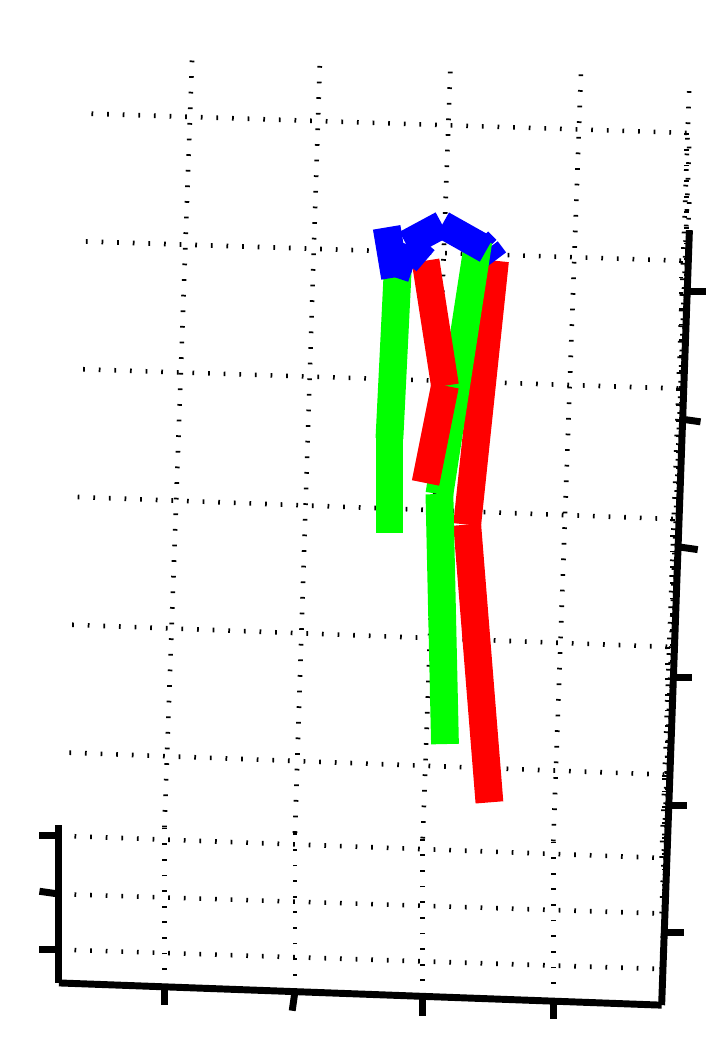}   
			&\includegraphics[width=0.12\linewidth, height=\mht]{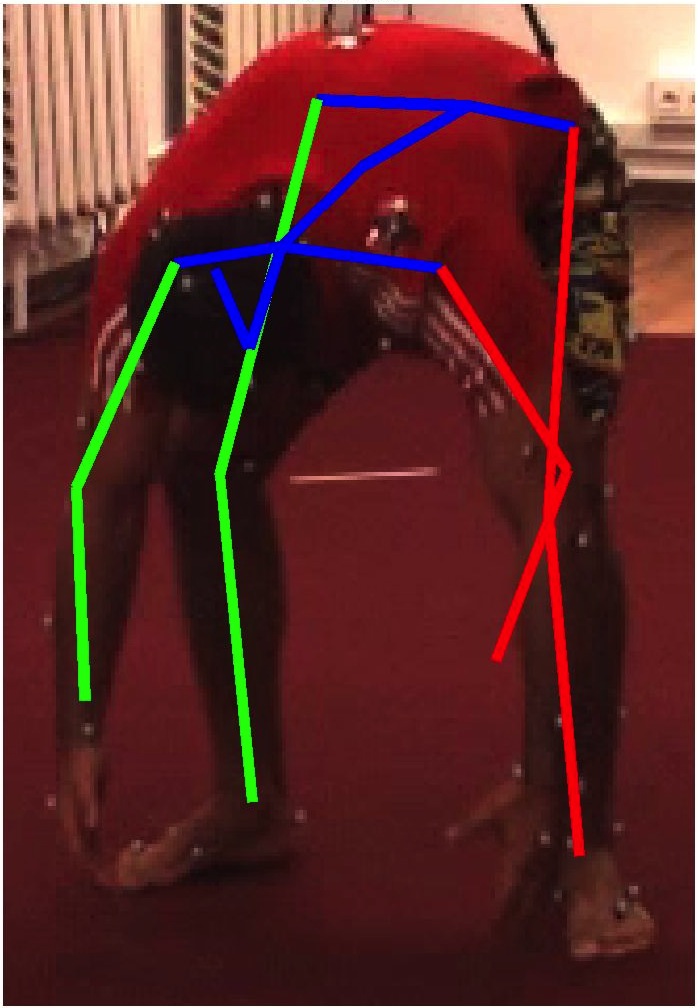}  
			&\includegraphics[width=0.13\linewidth, height=\mht]{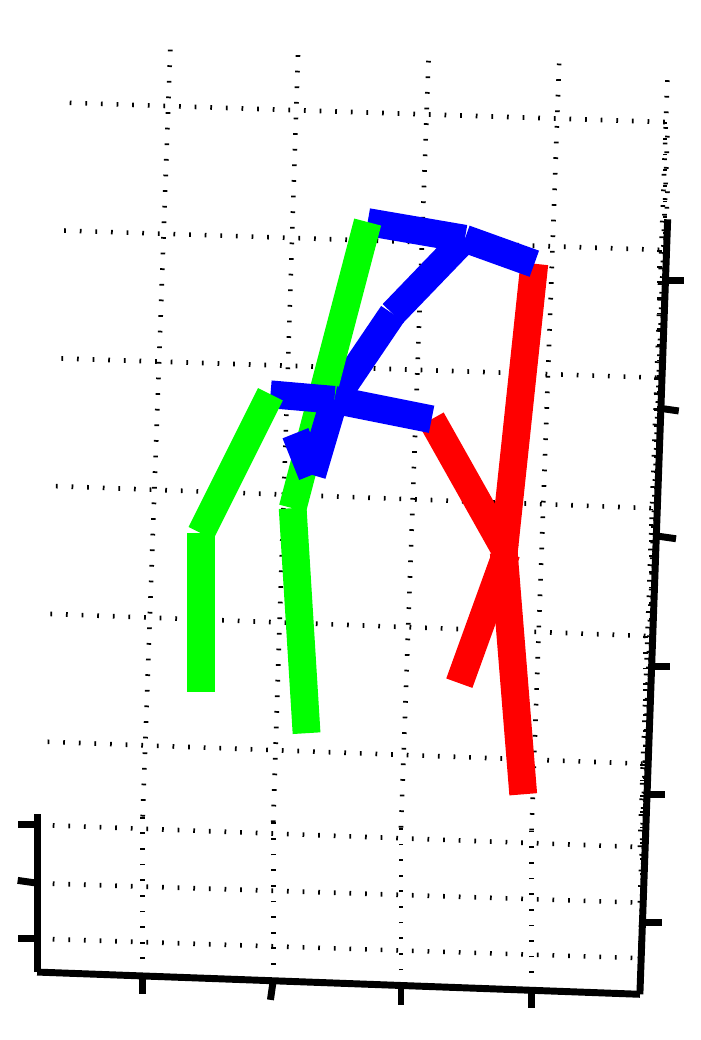} \\
			\includegraphics[width=0.12\linewidth, height=\mht]{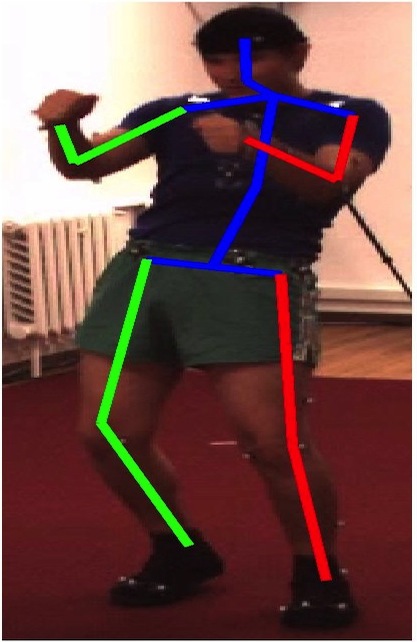} 
			&\includegraphics[width=0.115\linewidth, height=\mht]{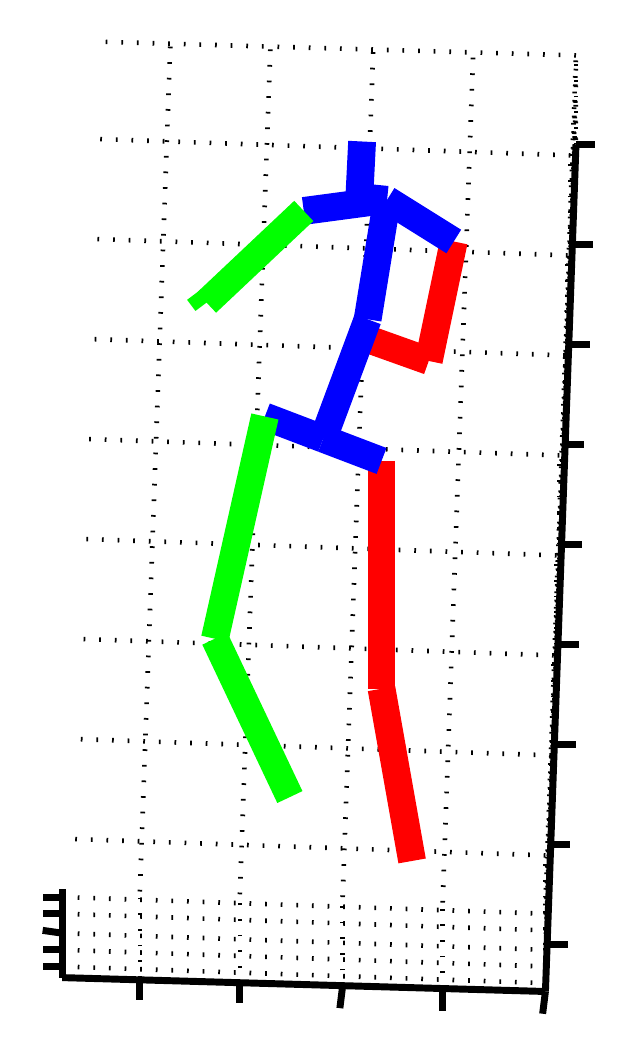}  
			&\includegraphics[width=0.12\linewidth, height=\mht]{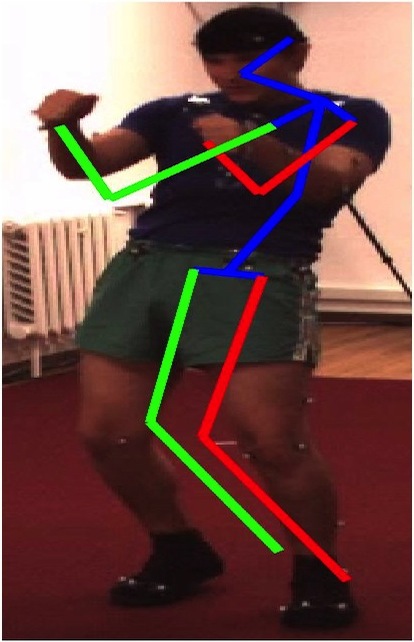}  
			&\includegraphics[width=0.115\linewidth, height=\mht]{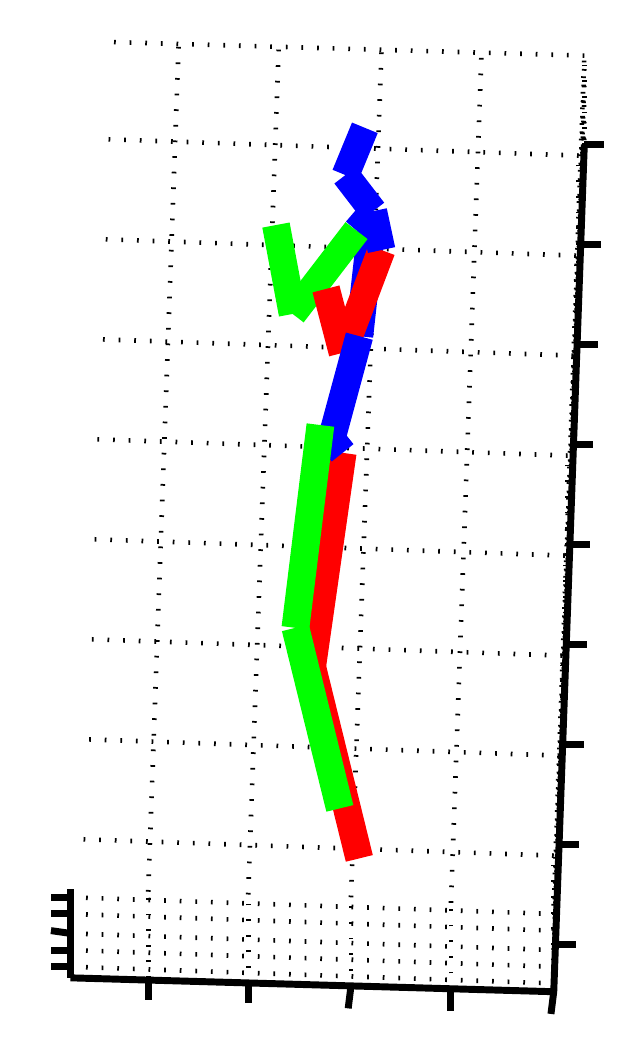}   
			&\includegraphics[width=0.12\linewidth, height=\mht]{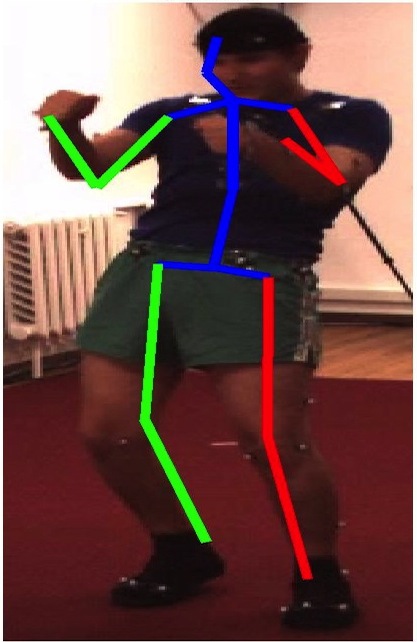}  
			&\includegraphics[width=0.115\linewidth, height=\mht]{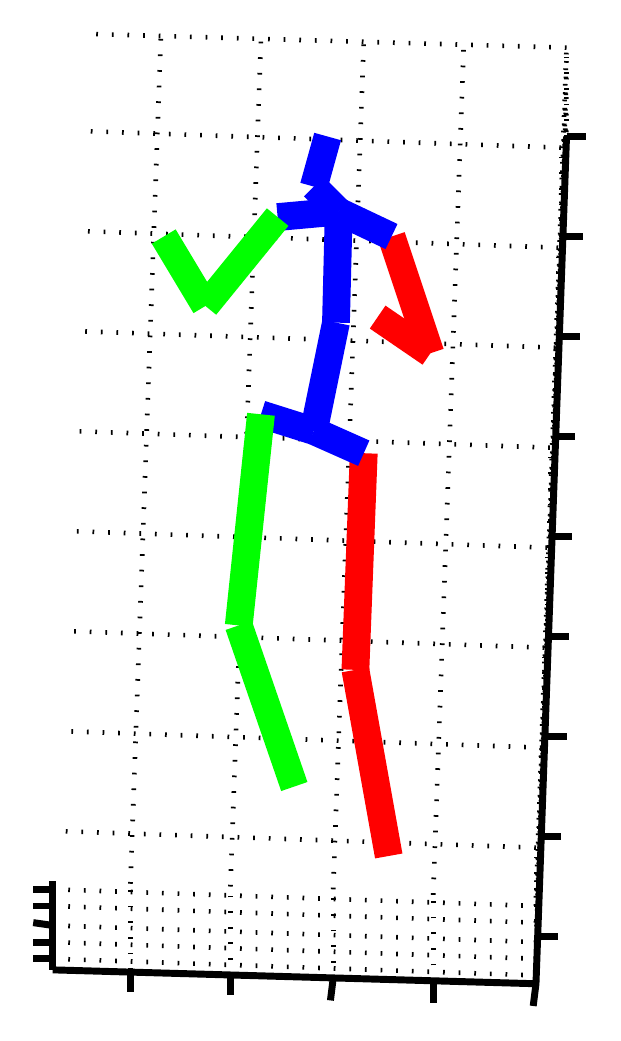} \\
			\includegraphics[width=0.12\linewidth, height=\mht]{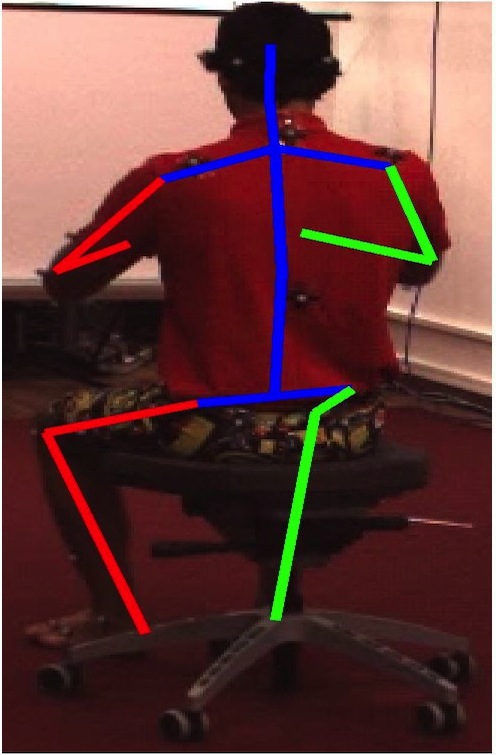} 
			&\includegraphics[width=0.13\linewidth, height=\mht]{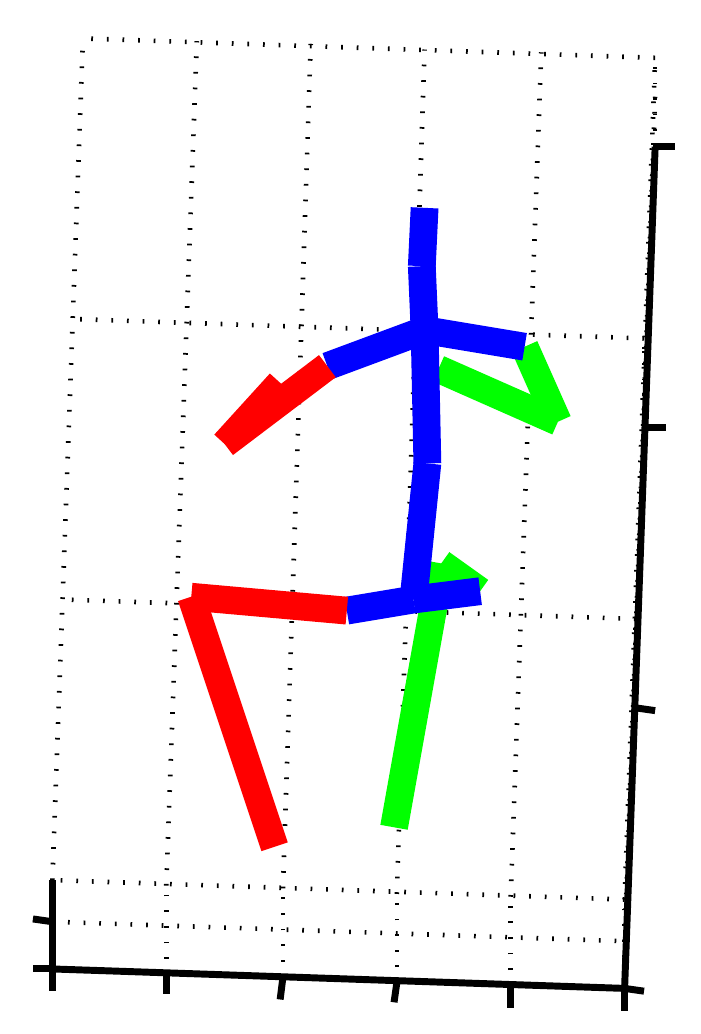}  
			&\includegraphics[width=0.12\linewidth, height=\mht]{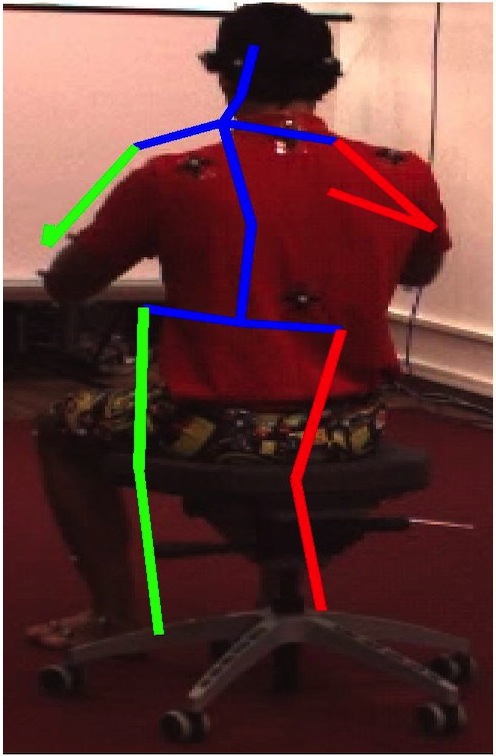}  
			&\includegraphics[width=0.13\linewidth, height=\mht]{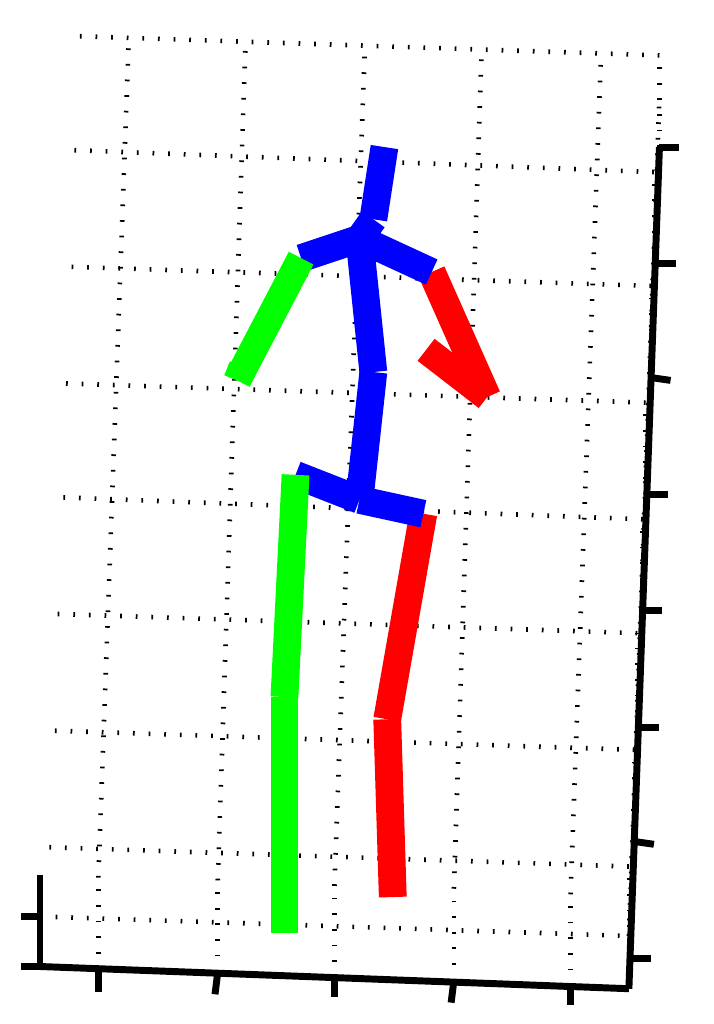}   
			&\includegraphics[width=0.12\linewidth, height=\mht]{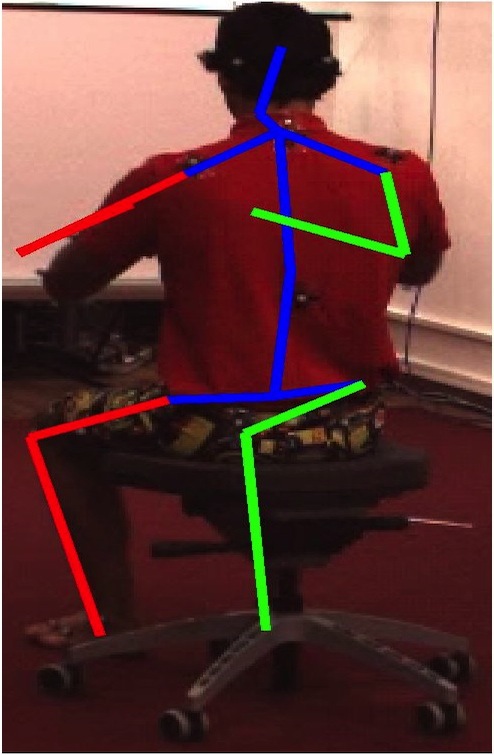}  
			&\includegraphics[width=0.13\linewidth, height=\mht]{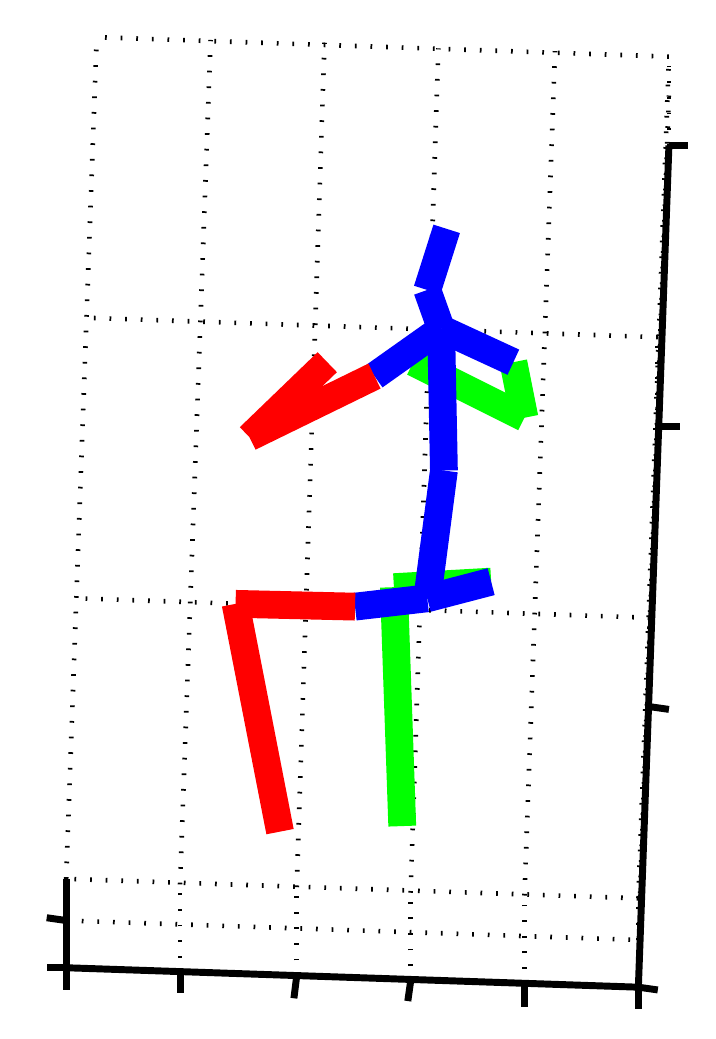}  \\
			\multicolumn{2}{c}{\footnotesize (a) Ground
				Truth}&\multicolumn{2}{c}{\footnotesize (b) Ionescu et.
				al~\cite{Ionescu14a}}&\multicolumn{2}{c}{\footnotesize (c) Our method}\\
		\end{tabular}
	}
	\caption{Pose  estimation results  on  Human3.6m.  The  rows  correspond to the
		\emph{Buying},  \emph{Discussion}  and  \emph{Eating}  actions. \textbf{(a)}
		Reprojection in the original images and  projection on the orthogonal plane of
		the  ground-truth  skeleton for  each  action.   \textbf{(b,c)} The skeletons
		recovered by  the approach of~\cite{Ionescu14a} and our method. Note that our
		method can  recover  the  3D  pose  in these  challenging  scenarios,  which
		involve significant amounts of self occlusion  and orientation ambiguity.}
	\label{fig:results}
\end{figure*}

Table~\ref{tab:overall}    summarizes    our    results\footnote{The sequence
corresponding   to  Subject   11  performing \emph{Directions} action on  camera
1 in trial 2 is  removed from evaluation due  to  video   corruption.} on
Human3.6m    and Figs.~\ref{fig:results}-\ref{fig:regressors}  depict some  of
them  on selected frames.   We  provide  additional   figures  in  the
supplementary material. Overall, our method significantly outperforms
$e^{\chi_2}$-HOG+KDE~\cite{Ionescu14a}   for all actions, with the mean error
reduced by about 23\%.  It also outperforms the method of~\cite{Ionescu14b},
which itself reports an overall performance improvement of 17\% over
$e^{\chi_2}$-HOG+KDE and 33\%  over plain HOG+KDE on a subset of the dataset
consisting of single images. Furthermore, it improves    on
CNN-Regression~\cite{Li14a}  by a  margin of more  than 5\% for all  the actions
for  which accuracy  numbers are reported. The improvement is particularly
marked for  actions such as \emph{Walking} and \emph{Eating}, which involve
substantial   amounts  of   predictable  motion. For \emph{Buying},
\emph{Sitting}  and  \emph{Sitting Down},  using  the structural information  of
the  human  body,  RSTV+KDE  yields  better  pose estimation accuracy. On 12 out
of 15 actions and in  average over all actions in the dataset, RSTV+DN yields
the best pose estimation accuracy.

In the following, we analyze the importance of motion compensation and of
	the influence of the temporal window size on pose estimation accuracy.
	Additional analyses can be found in the supplementary material. 

\begin{figure*}[t]
	\centering
	\scalebox{0.92}{
		\begin{tabular}{cccccccc}
			\includegraphics[width=0.12\linewidth, height=\mht]{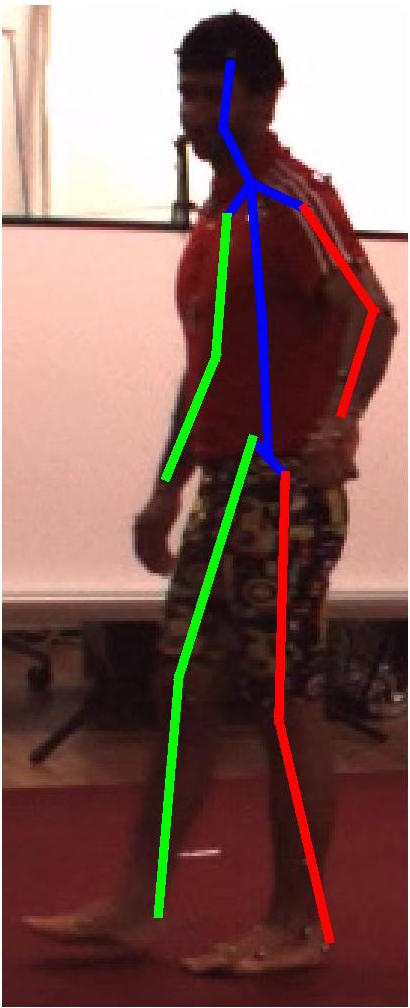} 
			&\includegraphics[width=0.11\linewidth, height=\mht]{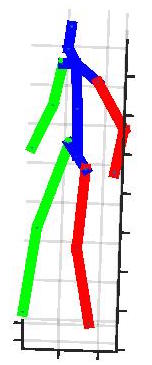}  
			&\includegraphics[width=0.12\linewidth, height=\mht]{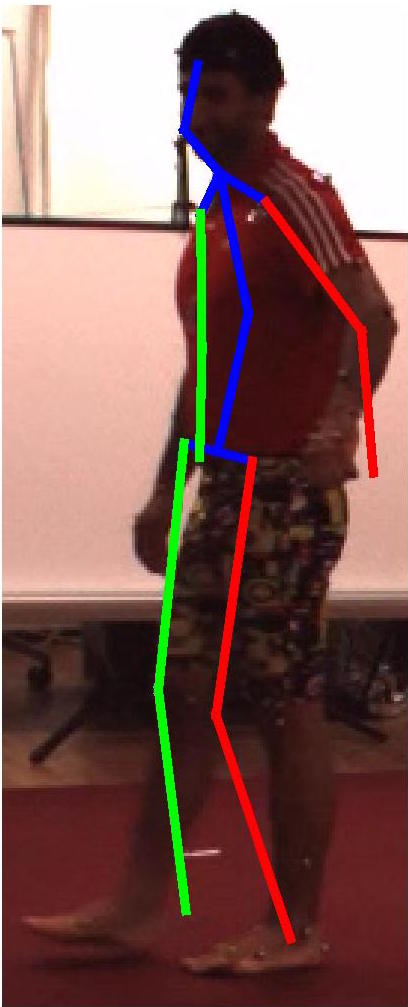}  
			&\includegraphics[width=0.11\linewidth, height=\mht]{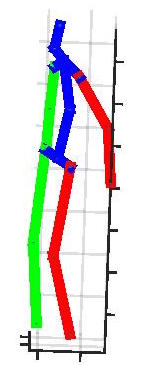} 
			&\includegraphics[width=0.12\linewidth, height=\mht]{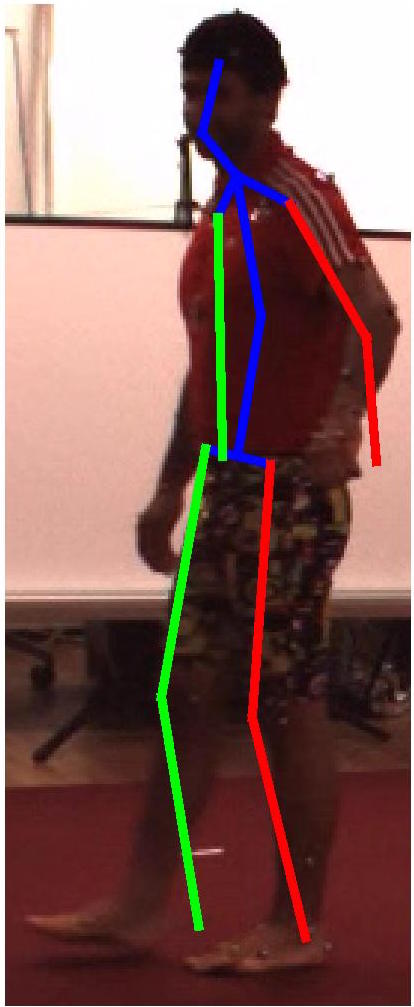}  
			&\includegraphics[width=0.11\linewidth, height=\mht]{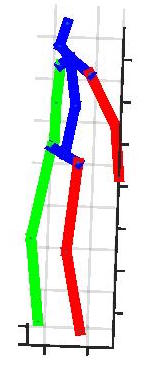}   
			& \includegraphics[width=0.12\linewidth, height=\mht]{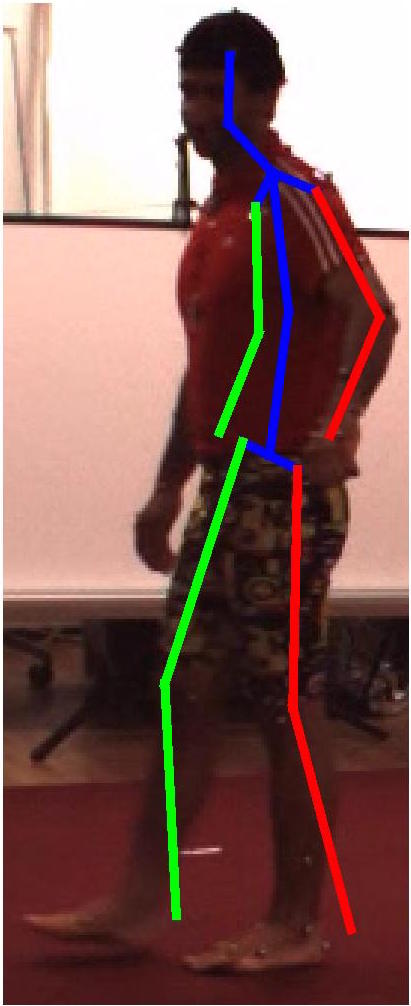} 
			&\includegraphics[width=0.11\linewidth, height=\mht]{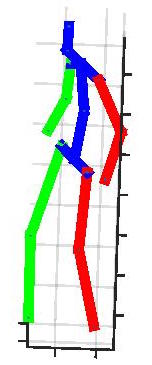}  \\
			\multicolumn{2}{c}{\footnotesize (a) Ground Truth}
			&\multicolumn{2}{c}{\footnotesize (b) RSTV+KRR}
			&\multicolumn{2}{c}{\footnotesize (c) RSTV+KDE}
			&\multicolumn{2}{c}{\footnotesize (d) RSTV+DN}\\
		\end{tabular}
	}
	\caption{3D  human  pose  estimation  with different  regressors  on Human3.6m.
		\textbf{(a)}  Reprojection  in  the  original images  and  projection  on the
		orthogonal plane of  the ground truth skeletons for  \emph{Walking Pair} action
		class.   \textbf{(b,c,d)}  The  3D body pose  recovered  using the KRR,  KDE or
		DN regressors applied to RSTV.}
	\vspace{-0.5cm}
	\label{fig:regressors}
\end{figure*}
\vspace{-4mm}

\paragraph{Importance of Motion Compensation.}
To highlight the  importance of motion compensation, we  recomputed our features
without it. We will refer to this  method as STV. We also tried using a recent
optical flow~(OF) algorithm for motion compensation~\cite{Park13}.

We provide results in Table~\ref{tab:motionstab} for two actions, which are
representative in the sense that the \emph{Walking Dog} one involves a lot of
movement while subjects performing the \emph{Greeting} action tend not to walk much.
Even without the motion compensation, regression on the features extracted from
spatiotemporal volumes yields better accuracy than the method
of~\cite{Ionescu14a}. Motion compensation significantly improves pose estimation
performance as compared to STVs. Furthermore, our CNN-based approach to motion
compensation (RSTV) yields higher accuracy than optical-flow based motion
compensation~\cite{Park13}.

\begin{table}[tbph]
	\begin{center}
		\setlength{\tabcolsep}{2pt}
		\scalebox{0.93}{
			\begin{tabular}[b]{lcccc}
				\toprule
				Action: 			&\cite{Ionescu14a}   &STV   & STV+OF~\cite{Park13} & RSTV \\
				\bottomrule
				\emph{Greeting} 	&164.39 		     &144.48& 140.97			   & {\bf127.12}\\
				\emph{Walking Dog}  &177.13  			 &138.66& 134.98			   & {\bf126.29}\\
				\bottomrule
			\end{tabular}
		}
	\end{center}
	\vspace{-3mm}
	
	\caption{Importance  of motion  compensation. The  results
	of~\cite{Ionescu14a} are compared against those of our method, without motion 
	compensation and with motion compensation using either optical flow~(OF)  
	of~\cite{Park13} or our algorithm introduced in
	Section~\ref{subsec:cnnmc}.}
	
	\vspace{-2mm}
	\label{tab:motionstab}  
\end{table}
\vspace{-4mm}

\paragraph{Influence of the Size of the Temporal Window.} In 
Table~\ref{tab:temporallength}, we  report the  effect of changing the size of
our temporal windows from 12 to 48 frames, again for two representative actions.
Using temporal information clearly helps 	and the best results  are obtained in
the range of  24 to 48 frames, which corresponds  to 0.5 to 1  second at 50 fps.
When the temporal window  is small,  the  amount of  information  encoded in the
features is not sufficient for accurate estimates. By contrast, with too large
windows, overfitting can be a problem as it becomes harder to account for
variation in the input data. Note that a temporal window size of 12 frames
already yields better results than the method of~\cite{Ionescu14a}. For  the
experiments we carried out  on Human3.6m, we  use $24$  frames as it  yields
both accurate reconstructions and efficient feature extraction.

\begin{table}[t]
	\begin{center}
		\setlength{\tabcolsep}{2pt}
		\scalebox{0.92}{
			\begin{tabular}[b]{l|c|ccccc}
				\toprule
				\multirow{2}{*}{Action:}	& \multirow{2}{*}{\cite{Ionescu14a}}  &
				\multicolumn{4}{c}{RSTV} \\
				&  				&12 frames&24 frames&36 frames&48 frames     \\
				\hline
				\emph{Walking}  &96.60 &58.78 &55.07 &{\bf53.68}&54.36		 \\
				\emph{Eating}   &132.37&93.97 &88.83 & 87.23	&{\bf85.36}    \\
				\bottomrule
			\end{tabular}
		}
	\end{center}
	\vspace{-3mm}
	\caption{Influence  of the size of the  temporal window.  We  compare the
		results of~\cite{Ionescu14a}  against those  obtained  using  our method,
		RSTV+DN, with increasing temporal window sizes.}
	\label{tab:temporallength}  
	\vspace{-3mm}
\end{table}

\subsection{Evaluation on HumanEva}

We further  evaluated our approach  on HumanEva-I and  HumanEva-II datasets. The
baselines we  considered are frame-based methods
of~\cite{Bo10,Elhayek15,Howe11,Kostrikov14,Simo-Serra12,Simo-Serra13,Wang14d},
frame-to-frame-tracking approaches which impose  dynamical priors on  the
motion~\cite{Sigal12,Taylor10b}  and the tracking-by-detection framework
of~\cite{Andriluka10}. The mean Euclidean distance between  the ground-truth 
and predicted  joint positions is  used to evaluate pose estimation performance. 
As the size of the training set in HumanEva is too small to train a deep 
network, we use RSTV+KDE instead of RSTV+DN.

We demonstrate  in Tables~\ref{tab:HumanEva}~and~\ref{tab:HumanEvaII}  that
using  temporal information  earlier in  the  inference process  in  a
discriminative  bottom-up fashion yields  more accurate results  than the
above-mentioned  approaches that enforce top-down temporal priors on the motion.

\vspace{-2mm}

\paragraph{HumanEva-I:}

\begin{figure}[t] \centering \scalebox{1}{
		\begin{tabular}{cccc}
			\includegraphics[width=0.7in, height=0.9in]{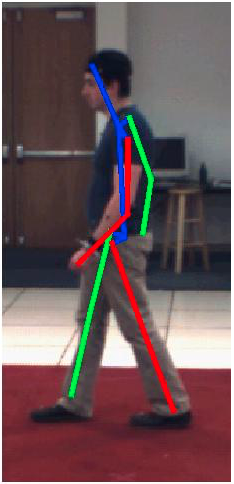}  \hspace{-3mm}
			&\includegraphics[width=0.7in, height=0.9in]{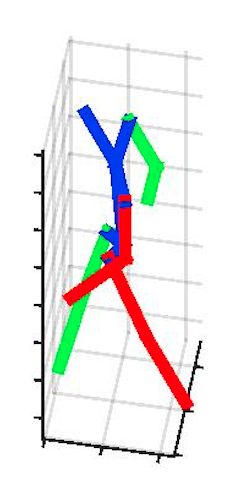}  \hspace{-3mm}
			&\includegraphics[width=0.7in, height=0.9in]{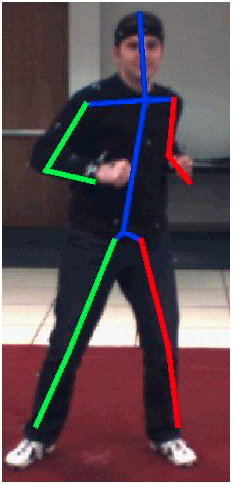} \hspace{-3mm}
			&\includegraphics[width=0.7in, height=0.9in]{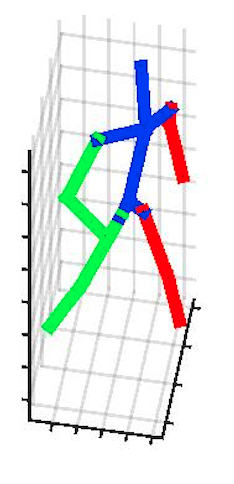}   \\
		\end{tabular}
	}
	\caption{Results on HumanEva-I.  The recovered 3D poses and their
		projection on the image are shown for \emph{Walking} and \emph{Boxing}
		actions. More results are provided in the supplementary material.}
	\vspace{-1mm}
	\label{fig:he1_results}
\end{figure}

For  the experiments  we carried  out on HumanEva-I, we  train our regressor on
training sequences of Subject $1$,  $2$ and $3$ and evaluate on the
``validation''  sequences  in  the  same  manner as  the  baselines  we  compare
against~\cite{Belagiannis14a,Bo10,Elhayek15,Kostrikov14,Sigal12,Simo-Serra13,Simo-Serra12,Taylor10b,Wang14d}.
Spatiotemporal features are computed only  from the first camera view. We report
the performance of our  approach on cyclic and  acyclic motions, more precisely 
\emph{Walking} and  \emph{Boxing}, in Table~\ref{tab:HumanEva} and depict
example 3D pose estimation results in Fig.~\ref{fig:he1_results}. The results 
show   that  our method   outperforms  the state-of-the-art approaches on this
benchmark as well.

\begin{table}[tbph]
	\vspace{-2mm}
	\centering
	\tabcolsep=0.1cm
	\scalebox{0.77}{
		\begin{tabular}[b]{lcccccccc}
			\toprule
			& \multicolumn{4}{c}{\emph{Walking}}   & \multicolumn{4}{c}{\emph{Boxing}}\\
			Method:					                & S1     & S2       & S3    & Avg.   & S1   	
			&
			S2   	 &  S3 	& Avg.	 \\
			\midrule
			Taylor et al.~\cite{Taylor10b}          & 48.8	 & 47.4   	& 49.8    & 48.7  
			& 75.35   & -	 & -	& -		  \\
			Sigal et al.~\cite{Sigal12}	            & 66.0   & 69.0   	& -       & -    
			& -       & -	 & -	& -		   \\
			Simo-Serra et al.'12~\cite{Simo-Serra12}& 99.6	 & 108.3    & 127.4   & 111.8
			& -       & -    & -    & -        \\
			Simo-Serra et al.'13~\cite{Simo-Serra13}& 65.1 	 & 48.6     & 73.5    & 62.2
			& -       & -    & -    & -         \\
			Wang et al.~\cite{Wang14d}			    & 71.9   & 75.7     & 85.3    & 77.6    & -
			& -    & -    & -         \\
			Belagiannis et al.~\cite{Belagiannis14a}& 68.3	 &  -       &  -      & -    
			& 62.70	  & -	 & -    & - 		\\
			Elhayek  et al.~\cite{Elhayek15}		& 66.5   &  -       & -       & -       &
			60.0    & -    & -    & -          \\
			Kostrikov et al.~\cite{Kostrikov14}	    & 44.0	 & 30.9     &{\bf41.7}&38.9  
			& -       & -    & -    & -           \\
			Bo et al.~\cite{Bo10}		     	    & 45.4	 & 28.3    	& 62.3    &45.33   
			&{\bf42.5}&64.0  &69.3	&58.6	         \\
			Ours			     			     	&{\bf37.5} &{\bf25.1} &49.2   &{\bf37.3}& 
			50.5   
			&{\bf61.7}&{\bf57.5}& {\bf56.6} \\
			\bottomrule
		\end{tabular}
	}
	\caption{3D joint position errors (in mm) on  the \emph{Walking}  and \emph{Boxing}  sequences 
	of HumanEva-I. We compare our approach  against methods that rely  on  discriminative 
	regression~\cite{Bo10,Kostrikov14},  2D pose detectors~\cite{Simo-Serra13,Simo-Serra12,Wang14d}, 3D  pictorial 
	structures~\cite{Belagiannis14a}, CNN-based markerless motion capture method 
	of~\cite{Elhayek15} and methods that rely on top-down temporal priors~\cite{Sigal12,Taylor10b}.
	`-' indicates that the results are not reported for the corresponding sequences.}
	\vspace{-4mm}
	\label{tab:HumanEva}  
\end{table}

\vspace{-3mm}

\paragraph{HumanEva-II:}
On HumanEva-II, we compare against~\cite{Andriluka10, Howe11} as they
report the best monocular pose estimation results on this dataset. HumanEva-II  provides only  a  
test  dataset  and no  training  data, therefore, we  trained our  regressors on  HumanEva-I using  
videos captured from different camera views. This demonstrates the generalization ability of our 
method to different camera views. Following~\cite{Andriluka10}, we  use subjects  $S1$, $S2$ and 
$S3$  from HumanEva-I  for training  and report  pose estimation  results in the first  $350$ 
frames of  the sequence featuring subject  S2. Global  3D joint positions in HumanEva-I are 
projected to camera coordinates for each view. Spatiotemporal features extracted from each camera 
view are mapped to 3D joint positions in its respective camera coordinate system, as done 
in~\cite{Poppe07}.  Whereas~\cite{Andriluka10}  uses  additional training data  from the 
``People''~\cite{Ramanan06b}  and ``Buffy''~\cite{Ferrari08} datasets, we  only use  the training 
data from  HumanEva-I. We evaluated our  approach using  the official  online  evaluation tool.   
We illustrate  the comparison in Table~\ref{tab:HumanEvaII}, where our method achieves the
state-of-the-art performance. 

\begin{table}[tbph]
	\centering
	\tabcolsep=0.1cm
	\scalebox{0.87}{
		\begin{tabular}[b]{lcccc}
			\toprule
			Method:					            & S2/C1 	   & S2/C2       & S2/C3       	 &Average	\\
			\midrule
			Andriluka et al.~\cite{Andriluka10} & 107	   	   & 101         & -		     &-         \\
			Howe~\cite{Howe11}			 		& 81		   & {\bf 73}	 & 143           &99        \\
			Ours			     	     		&{\bf79.6}     &79.0         & {\bf79.2}	 &{\bf79.3} \\
			\bottomrule
		\end{tabular}
	}
	\caption{3D joint position errors (in mm) on the {Combo} sequence of the
		HumanEva-II dataset. We compare  our  approach  against  the  tracking-by-detection  
		framework of~\cite{Andriluka10} and recognition-based method of~\cite{Howe11}. `-' 
		indicates that the result is not reported for the corresponding sequence.}
	\vspace{-3mm}
	\label{tab:HumanEvaII}  
\end{table}

\subsection{Evaluation on KTH Multiview Football Dataset}
\label{subsec:kth_eval}

\begin{figure}[t]
	\centering
	\scalebox{0.92}{
		\begin{tabular}{ccc}
			\includegraphics[width=0.9in]{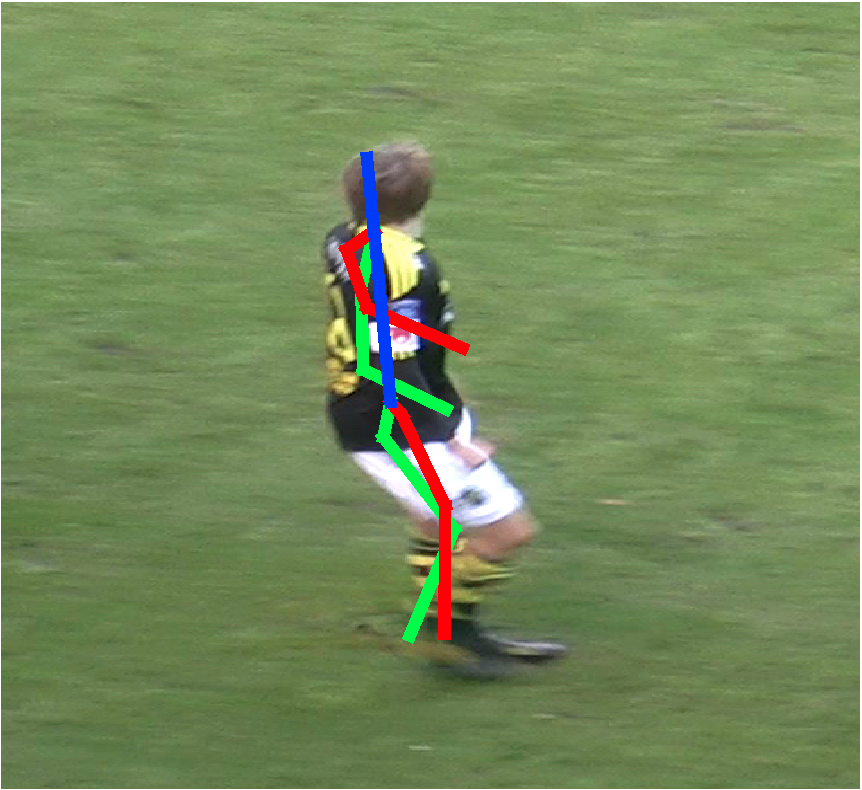} 
			&\includegraphics[width=0.9in]{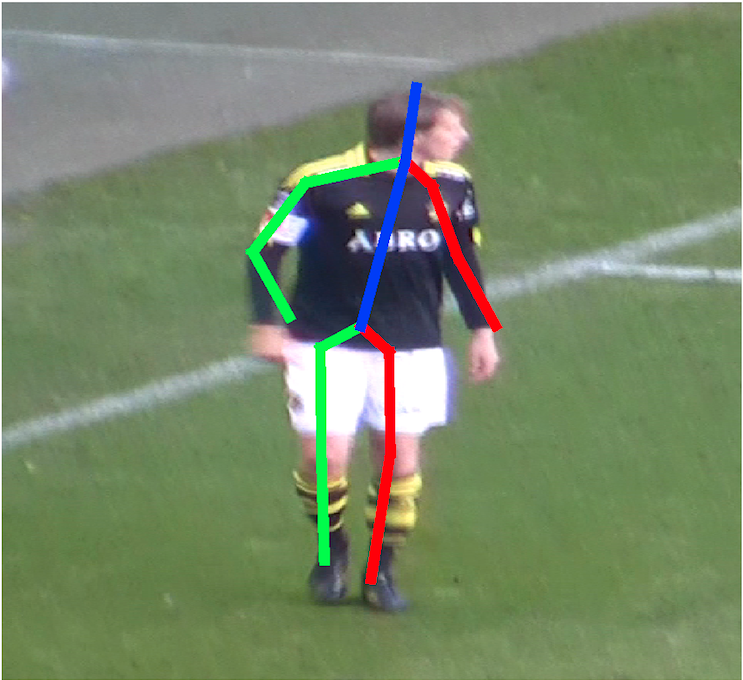}  
			&\includegraphics[width=0.9in]{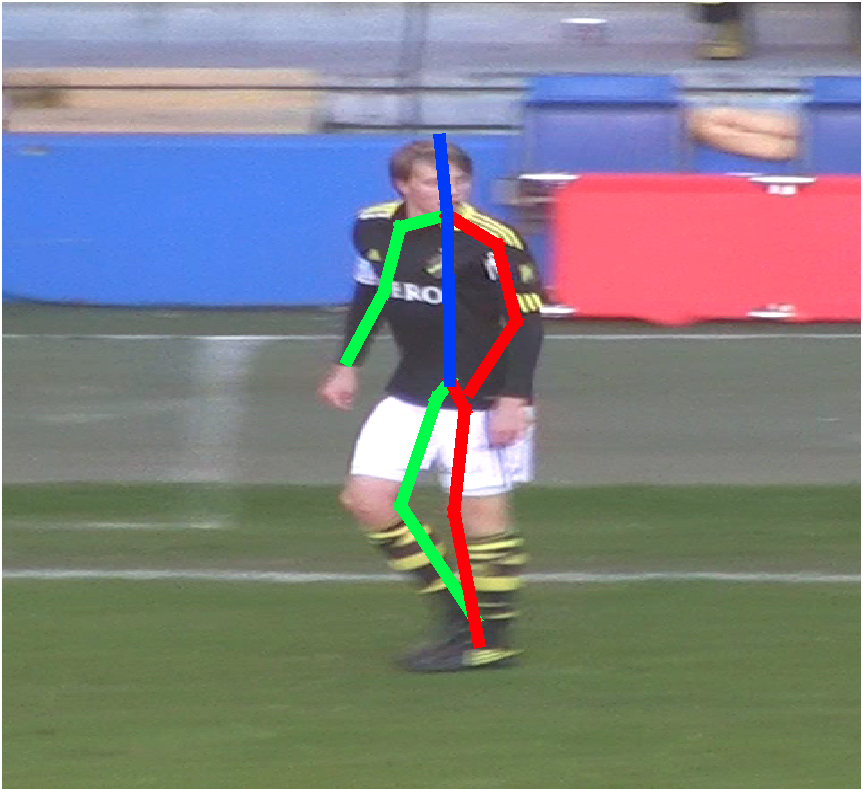}   \\
			\footnotesize Camera 1 & \footnotesize Camera 2 &  \footnotesize Camera 3
			\\
		\end{tabular}
	}
	\caption{Results on KTH Multiview Football II. The 3D skeletons are recovered from Camera 1 
	images and projected on those of Camera 2 and 3, which were not used to compute the poses.}
	\vspace{-2mm}
	\label{fig:kth_results}
\end{figure}

As in~\cite{Belagiannis14a,Burenius13},  we evaluate  our method  on the sequence containing  
Player 2.   The first  half of the  sequence is  used for 	training  and  the  second  half  
for   testing,  as  in  the  original work~\cite{Burenius13}. To compare our results to those 
of~\cite{Belagiannis14a,Burenius13}, we report pose estimation accuracy in terms of percentage of 
correctly estimated 	parts (PCP) score. As in the HumanEva experiments, 	we provide results for 
RSTV+KDE. Fig.~\ref{fig:kth_results} depicts 	example pose estimation results. As shown  in 
Table~\ref{tab:kth},  we outperform  the baselines  even  though our  algorithm is  monocular,  
whereas they  use both cameras.  This is due  to the  fact that  the baselines  instantiate 3D 	
pictorial structures relying on 2D  body part detectors, which  may not be precise when the 
appearance-based information is weak. By contrast,	collecting appearance and motion information 
simultaneously from rectified spatiotemporal volumes, we achieve better 3D pose estimation accuracy.

\begin{table}[t]
	\centering
	\tabcolsep=0.1cm
	\scalebox{0.87}{
		\begin{tabular}[b]{lcccc}
			\toprule
			Body parts: &\cite{Burenius13} (Mono)   &\cite{Burenius13} (Stereo) &\cite{Belagiannis14a}(Stereo)  & Ours (Mono) \\
			\midrule
			Pelvis  	  &97      	&97			&-		&{\bf99}   \\
			Torso		  &87	    &90 		&-		&{\bf100}  \\
			Upper arms    &14 	    &53  		&64     &{\bf74}   \\
			Lower arms    &06		&28			&{\bf50}&49        \\
			Upper legs    &63		&88			&75 	&{\bf98}   \\
			Lower legs    &41		&{\bf82}	&66 	&77	       \\
			All parts     &43		&69			&-		&{\bf79}   \\
			\bottomrule
		\end{tabular}} 
		\caption{On the KTH Multiview Football II we have compared our method using a single camera 
		to those of~\cite{Burenius13} using either single or two cameras and to the one 
		of~\cite{Belagiannis14a} using two cameras. `-' indicates that the result is not reported 
		for the corresponding body part.}
		\vspace{-6mm}
		\label{tab:kth}  
\end{table}	
\vspace{-1mm}

\vspace{-1mm}
\section{Conclusion}
\vspace{-1mm}
We have demonstrated  that taking into account motion information  very early in the modeling 
process  yields significant performance improvements  over doing it {\it a  posteriori} by linking  
pose estimates in  individual frames. We have shown that extracting appearance and motion cues from 
rectified spatiotemporal volumes disambiguate challenging  poses with mirroring and self-occlusion, 
which brings about substantial increase in accuracy over the state-of-the-art methods on several 3D 
human pose estimation benchmarks. Our proposed framework is generic and could be used for other 
kinds of articulated motions.

\vspace{2mm}\noindent {\bf Acknowledgments.} This work was supported in part by the EUROSTARS Project CLASS.

{\small
\bibliographystyle{ieee}
\bibliography{string,short,vision,learning,biomed}
}

\end{document}